\documentclass[lettersize,journal,twoside]{IEEEtran}
\usepackage{amsmath,amsfonts}
\usepackage{algorithmic}
\usepackage{algorithm}
\usepackage{array}
\usepackage[caption=false,font=normalsize,labelfont=sf,textfont=sf]{subfig}
\usepackage{textcomp}
\usepackage{stfloats}
\usepackage{url}
\usepackage{verbatim}
\usepackage{graphicx}
\usepackage{cite}
\usepackage{multirow}
\usepackage{graphicx}

\usepackage[colorlinks,
            linkcolor=red,
            anchorcolor=blue,
            citecolor=green
            ]{hyperref}

\usepackage{subcaption}  
\usepackage{caption}
\usepackage{array}  
\usepackage{hyperref}
\usepackage{array}  
\usepackage{xcolor} 

\usepackage{algorithm}
\usepackage{algorithmic}
\usepackage{float}

\usepackage{makecell}

\usepackage{booktabs}
\usepackage{hhline}

\usepackage[table]{xcolor}
\usepackage{colortbl}
\usepackage{amssymb}
\usepackage[utf8]{inputenc}
\usepackage{tabularx}

\usepackage{xcolor}

\usepackage[authormarkup=none, commandnameprefix=always]{changes}
\definechangesauthor[name={Reviewer}, color=red]{R}

\usepackage{fancyhdr}
\newcommand{\myarticleno}{XXXX} 
\newcommand{\myjournal}{IEEE TRANSACTIONS ON MULTIMEDIA, VOL. XX, XXXX} 
\newcommand{\myauthor}{ROY} 
\newcommand{\mytitle}{Vision-Language Guided Hyperspectral Object Tracking} 

\fancypagestyle{plain}{%
  \fancyhf{}%
  \fancyhead[LO]{\footnotesize\myarticleno}%
  \fancyhead[RO]{\footnotesize\myjournal}%
  \fancyfoot[C]{\thepage}%
}

\begin{document}

\title{Vision-Language Guided Hyperspectral Object Tracking via Semantics Fusion and Contextual Template Updating}

\author{Rui~Yao,
        Yuhong~Zhang,
        Kunyang~Sun,
        Hancheng~Zhu,
        Jiaqi~Zhao,~\IEEEmembership{Member,~IEEE},
        Zhiwen~Shao,
        and Abdulmotaleb~El~Saddik,~\IEEEmembership{Fellow,~IEEE}
\thanks{This work was supported in part by the National Natural Science Foundation of China under Grant 62172417 and 62472424. \emph{(Corresponding authors: Kunyang~Sun.)}}
\thanks{Rui~Yao, Yuhong~Zhang, Kunyang~Sun, Hancheng~Zhu, Jiaqi~Zhao and Zhiwen~Shao are with the School of Computer Science and Technology / School of Artificial Intelligence, and Mine Digitization Engineering Research Center of the Ministry of Education, and Jiangsu Provincial Industrial Technology Engineering Center for Intelligent Sensing and Emergency IoT in Underground Space, China University of Mining and Technology, Xuzhou 221116, China (E-mail: \{ruiyao, ts24170043a31ld, kunyang\_sun, zhuhancheng, jiaqizhao, zhiwen\_shao\}@cumt.edu.cn). }
\thanks{Abdulmotaleb El Saddik is with the School of Electrical Engineering and Computer Science, University of Ottawa, Ottawa, ON K1N 6N5, Canada (E-mail: elsaddik@uottawa.ca).}}


\maketitle
\thispagestyle{fancy}

\begin{abstract}
Hyperspectral object tracking (HOT) leverages the rich spectral information provided by hyperspectral videos (HSVs), offering substantial potential for object tracking. However, efficiently extracting and exploiting spectral information from redundant spectral bands remains a fundamental challenge, which severely limits model generalization and tracking performance. Moreover, in dynamic scenes, targets often experience drastic appearance variations due to factors such as occlusion and illumination changes.
These variations lead to large deformations between the current frame and the template. Such discrepancies pose major challenges for existing temporal modeling approaches. In this work, we propose VLHTrack, a novel hyperspectral vision-language (VL) joint tracking framework.  
Specifically, we incorporate language priors to address the fundamental challenge of spectral redundancy by designing a Language-Guided Band Selection Module (LBSM). By leveraging Large Language Model (LLM) descriptions, LBSM establishes a semantic-to-spectral mapping that mitigates redundancy and accentuates discriminative spectral features. A Multi-Modal Vision–Language Fusion Module is then employed to seamlessly integrate visual and linguistic embeddings, harnessing their complementary advantages to learn coherent cross-modal representations. To address target deformation in long-term sequences, we propose a dynamic update template feature strategy implemented via the Dynamic Template Update with Mamba (DTUM) module. By leveraging selective state space modeling, DTUM learns inter-frame dependencies to update template feature, ensuring efficient template feature evolution guided by temporal context. Experiments on HOT2023 and HOT2024 demonstrate that VLHTrack outperforms state-of-the-art (SOTA) methods. The source code is available at \href{https://github.com/rayyao/VLHTrack}{https://github.com/rayyao/VLHTrack}.

\end{abstract}


\section{Introduction}
\label{sec:intro}

\IEEEPARstart{V}{isual} object tracking (VOT) is a fundamental task in computer vision with broad applications in face tracking \cite{9199576}, autonomous driving \cite{9693129, yao2025adversarial}, robotics \cite{9127813}, and medical imaging. The goal is to estimate the spatial location of a target across frames given its initial state \cite{6671560}. However, conventional RGB-based trackers are limited by their reliance on only three channels, making them vulnerable to variations in illumination, scale, pose, occlusion, background clutter, and interference from similar objects \cite{wang2025deep}. These challenges are particularly evident when targets share similar visual characteristics with their surroundings, often leading to tracking failures \cite{wang2025ssf}. Recent advances in hyperspectral snapshot sensors have enabled the acquisition of HSVs \cite{5543780}, which contain rich spatial, spectral, and temporal information.
Unlike RGB imagery, hyperspectral data provide continuous spectral bands that enhance material discrimination and enable more precise target localization even in cluttered environments \cite{gewali2018machine}.

Despite its considerable potential, HOT remains constrained by the high dimensionality and inter-band redundancy of hyperspectral images. This limitation severely constrains feature representation and increases model sensitivity to spectral noise and inter-band variations, reducing the stability and generalization of tracking models. Early correlation filter–based (CF) methods \cite{6870486, 9924160, 8960632} primarily relied on handcrafted spectral–spatial features and selected bands using statistical criteria such as mutual information, inter-class variance, and inter-band correlation.
 However, their limited representational capacity leads to degraded performance under complex backgrounds and illumination changes. 
 Recent deep learning–based (DL) trackers \cite{10798510, zhang2025historical, chen2025ssttrack, 10128966, 10438474, 10149343} introduce mechanisms such as band attention, spectral unmixing, and prompt learning to achieve adaptive band modeling, improving feature discriminability and cross-modal fusion. 
Nevertheless, most still rely on fixed or heuristic band selection, making it difficult to adaptively choose suitable spectral bands under varying scene conditions.

Recently, studies on VL tracking models \cite{10659157, 10149530, guo2022divert, feng2021siamese} have shown that incorporating textual priors can significantly enhance visual understanding and representation learning. By integrating linguistic semantics with visual features, they provide additional cues such as object category, material, and color. This cross-modal integration leads to more discriminative and robust representations under appearance variations and occlusions. 
Although significant progress has been made in the fusion of vision and language for RGB tracking, and recently emerged HOT methods such as CCTrack\cite{wang2025hyperspectral} have begun to leverage class consistency, the potential of natural language for extracting spectral information remains underutilized. The challenge in transitioning from RGB images to hyperspectral images lies in how to efficiently utilize high-dimensional spectral information.
Inspired by these studies, we posit that language priors can guide spectral band selection by introducing scene-level semantic understanding into the model, thereby enabling more adaptive and semantically grounded spectral feature modeling. 

\begin{figure}[!t] 
    \centering
    \includegraphics[width=\columnwidth]{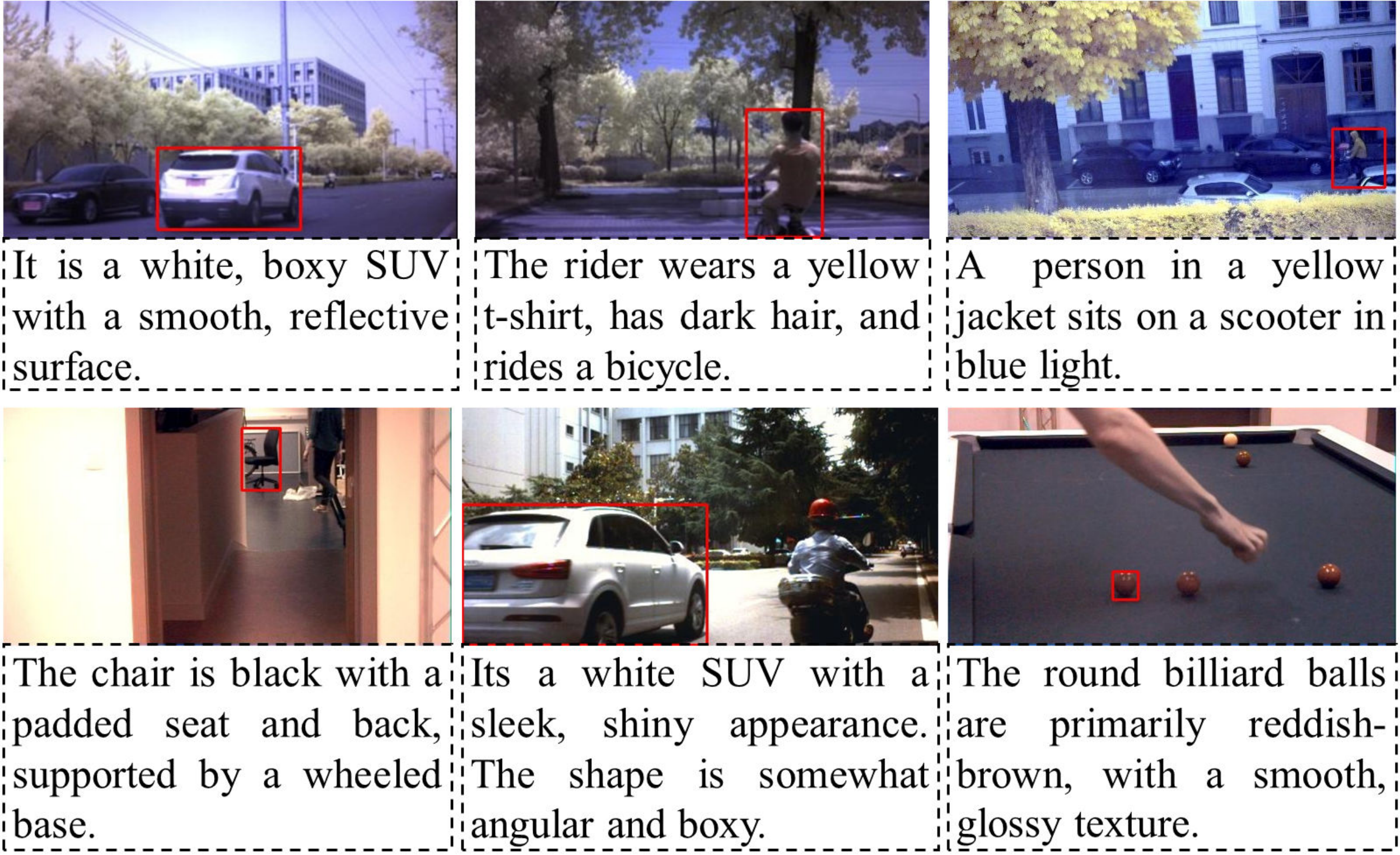} 
    \caption{{
    Representative HSV sequences with LLM-generated language descriptions. False-color initial frames highlight annotated target regions. These descriptors serve as stable semantic anchors for band selection and multi-modal interaction.
    }}
    \label{fig:language_description}
\end{figure}

To achieve this goal, we propose the vision–language framework for HOT. It leverages LLMs \cite{achiam2023gpt} to generate natural language descriptions that supply semantic priors for tracking. 
Existing trackers often lack high-level semantic guidance, struggling to distinguish targets from cluttered backgrounds. To bridge this gap, we leverage language priors as a discriminative constraint. As illustrated in Fig.~\ref{fig:language_description}, by generating descriptive attributes (e.g., ``reflective surface'' or ``yellow jacket'') from the initial frame, our framework establishes a stable semantic anchor. These linguistic cues equip the model with an explicit understanding of the target, enabling the subsequent LBSM to prioritize spectral bands that physically align with the target's inherent properties.
To effectively exploit spectral information, LBSM also combines entropy analysis and redundancy modeling to eliminate inefficient spectral bands while retaining the most discriminative ones. 
Unlike traditional methods that rely solely on statistical correlations for band selection and lack target-awareness, LBSM aligns hyperspectral and linguistic features in a shared semantic embedding space, thereby achieving semantically consistent band selection. This guides the model to focus on spectral bands physically relevant to the target's attributes. Unlike traditional data-driven methods or implicit attention mechanisms that rely on unsupervised statistical correlations, LBSM functions as a knowledge-driven system. By grounding the selection process in high-level linguistic semantics, it transforms band selection from an unsupervised correlation search into a cross-modal supervised mapping. This ensures the selected bands are not only statistically informative but also physically consistent with the target's intrinsic attributes.
Notably, linguistic information contributes not only to band selection but also imposes semantic constraints during subsequent tracking, significantly enhancing representation learning and tracking accuracy.

{Although deep learning has significantly advanced HOT, most existing trackers rely primarily on superficial statistical correlations for band selection, failing to establish a semantic-to-spectral mapping that bridges physical attributes and spectral signatures. Consequently, lacking semantic understanding, these data-driven approaches are highly vulnerable to spectral confusion. This is especially problematic when background distractors exhibit similar spectral responses but differ fundamentally in semantic identity.}

Furthermore, in long-term tracking, occlusions and illumination variations often cause significant discrepancies between the target's current appearance and the initial template. Consequently, relying solely on a fixed initial template hinders the construction of robust temporal models.
To address this, recent works such as HyMamba \cite{gao2025hyperspectralmambahyperspectralobject} have pioneered the use of state space models to capture long-range temporal dependencies. However, while effective, current approaches primarily rely on visual feature updates, often overlooking the stability provided by semantic information. 
Moreover, these challenges often cause significant discrepancies between the target and the initial template feature, and relying solely on the initial template feature makes it difficult to construct reliable temporal modeling. Then we argue that the template feature should be dynamically updated throughout tracking to better adapt to complex and evolving appearance changes \cite{chen2025improving, 10375560}. Accordingly, we propose a dynamic template feature update strategy, termed DTUM, which employs selective state space modeling to capture inter-frame dependencies and dynamically update template feature, ensuring temporal coherence in target representation. 

The contributions of this work are summarized as follows:
\begin{itemize}
    \item We propose VLHTrack, a task-specific vision-language framework tailored to address the spectral challenges in HOT, which leverages LLM-generated textual descriptions to construct an interpretable mapping between the semantic and spectral–physical spaces. Extensive experiments demonstrate that VLHTrack achieves superior performance on HOT2023 and HOT2024 datasets.
    \item We design LBSM, which integrates entropy, structural redundancy, and semantic similarity guided by language embeddings to select discriminative band combinations and facilitate language-informed spectral feature learning.
    \item We develop DTUM, a selective state-space-based temporal modeling module designed to dynamically update template feature by capturing inter-frame dependencies, thereby preserving temporal coherence and enhancing overall tracking stability. 
\end{itemize}

\section{Related Work}
\subsection{Hyperspectral Object Tracking}
Hyperspectral object tracking methods can be broadly divided into CF-based approaches and DL-based approaches. CF-based methods rely on handcrafted spectral–spatial features to distinguish targets from the background. Early correlation-filter methods, such as KCF \cite{6870486}, relied on circulant matrices and Fourier transforms for efficient training. Later extensions, including MHT \cite{8960632}, CNHT \cite{qian2018object}, and TSCFW \cite{9924160}, incorporated spectral–spatial priors and regularization to enhance robustness.
Although CF-based approaches are computationally efficient, their reliance on shallow handcrafted features limits robustness under large-scale appearance variations and long-term occlusion. 

In contrast, DL-based methods employ end-to-end networks to automatically learn discriminative spectral–spatial representations, demonstrating stronger adaptability in complex environments. SiamHYPER \cite{9933370} and SiamBAG \cite{10149343} transfer pretrained RGB models into the hyperspectral domain via cross-modal attention and band-grouping, respectively. However, fixed grouping strategies limit adaptability when spectral bands vary. 
To overcome this, BAE-Net \cite{9191105} and Hy-Tracker \cite{10569013} introduced attention-based band selection to dynamically identify discriminative subsets. Transformer-based models, such as 
TMTNet \cite{zhao2023tmtnet}, TFTN \cite{9924197}, and HTACPE \cite{10820018}, advanced multi-modal 
fusion and alleviated overfitting, but still rely on pixel-level registration with RGB data.
More recent works, including HDSP \cite{10798510} and ProFiT \cite{chen2025profit}, enhance cross-modal adaptability via prompt learning and frequency-domain modeling, while SSTtrack \cite{chen2025ssttrack} and HOPL \cite{zhang2025historical} improve temporal robustness through generative modeling and history-aware prompting. For efficiency, SiamOHOT \cite{10225562} applies 
knowledge distillation, HotMoE \cite{sun2025hotmoe} adopts a sparse Mixture-of-Experts framework, and MMF-Net \cite{10438474} aligns spectral features with physical material attributes under occlusion. 
Despite these advances, hyperspectral trackers still face challenges in spectral adaptability, temporal modeling, and cross-modal alignment. 
In addition to traditional statistical methods, learnable band selection has gained significant attention. Recent works such as BCR-Net \cite{rs17223689} focus on joint band and context refinement, and supervised embedded methods \cite{zimmer2025supervisedembeddedmethodshyperspectral} integrate band pruning directly into the feature embedding process. 
Recently, CCTrack \cite{wang2025hyperspectral} leveraged inter-frame category consistency to stabilize tracking. However, such coarse label-based approaches lack the granularity required to handle diverse environmental interferences. Distinct from these, VLHTrack introduces a semantic-to-spectral mapping that goes beyond simple classification labels, aligning high-level semantics directly with low-level spectral-physical properties to enable target-specific band selection.

Furthermore, traditional band selection paradigms are inherently data-driven and rely on internal statistical regularities, which often fail to distinguish the target from semantically irrelevant background clutter. While attention mechanisms assign weights across all bands and frequently preserve redundant noise, LBSM utilizes natural language as an external supervision signal. This synergy of entropy-based filtering and semantic-guided clustering achieves a sparse selection, providing much stronger constraints for spectral redundancy reduction than implicit or self-supervised weighting methods.

\subsection{Vision-Language Tracking}
Vision–Language (VL) tracking \cite{10659157, 10149530, guo2022divert, feng2021siamese} integrates natural language descriptions with visual inputs to enhance target discrimination, reduce ambiguities, and improve generalization across diverse scenes. 
SNLT \cite{feng2021siamese} fused static word embeddings with Siamese features but lacked contextual modeling. 
Later, CTVLT \cite{feng2025enhancing} introduced zero-shot language grounding via Grounding DINO for open-vocabulary localization, though temporal consistency remained insufficient. 
With the rise of large VL models, research has shifted toward parameter-efficient and scalable cross-modal modulation paradigms. 
CPIPTrack \cite{10817474} employs CLIP \cite{radford2021learning} as a frozen backbone using interactive prompts, improving semantic guidance while reducing parameters. MemVLT \cite{feng2024memvlt} leverages memory-augmented prompts for long-term robustness, while DTVLT \cite{li2024dtvlt} analyzed the role of description update frequency in temporal consistency. VLTVerse \cite{li2024texts} explored the complementarity of multi-granularity language inputs, CiteTracker \cite{10377155} introduced adaptive text weighting to mitigate noisy descriptions, and MAVLT \cite{10948517} adopted state space modeling for efficient cross-modal fusion. 
Furthermore, while text-guided learning has been explored to enhance category discriminability in static hyperspectral image classification (e.g., LDGnet \cite{zhang2023language}, TASA-Net \cite{wang2025text}), its application in dynamic tracking remains scarce. In particular, few methods attempt to establish a mapping from textual descriptions to spectral band selection, leaving this connection largely unexplored. Distinct from these works, our VLHTrack does not merely use language for spatial prompting. Instead, it addresses the unique hyperspectral challenge of band redundancy by establishing a mapping between high-level semantics and low-level spectral-physical properties. This allows the model to select discriminative bands based on the physical characteristics of the target class.

\subsection{State Space Models}
State Space Models (SSMs) \cite{gu2021efficiently} have recently emerged as a highly efficient and scalable paradigm for temporal sequence modeling.
The early structured S4 model \cite{gu2022parameterization} enabled long-range dependency modeling through linear time-invariant systems but was limited under dynamic conditions. Mamba \cite{gu2023mamba} addressed this by introducing selective input-dependent parameters, enhancing adaptability and efficiency, and making it well-suited for template updating in visual tracking. Extensions such as VideoMamba \cite{li2024videomamba}, LocalMamba \cite{huang2024localmamba}, and  VMamba \cite{liu2024vmamba} expanded applications to video and multi-object tracking.
Despite these advances, existing SSM-based trackers remain largely focused on RGB videos, with limited exploration of spectral redundancy, material-specific attributes, and multi-modal coupling in hyperspectral scenarios.
{Recently, the emergence of SSMs has opened new avenues for HOT. For instance, HyMamba \cite{gao2025hyperspectralmambahyperspectralobject} establishes a unified framework for efficient spectral-temporal dependency modeling, while the Prompt-based Mamba \cite{10876425} utilizes spectral-temporal tokens to guide adaptive feature learning. Our VLHTrack differs from these works by incorporating linguistic semantics as a guide to prune spectral redundancy before employing state-space modeling.}

\begin{figure*}[!t]
    \centering
    \includegraphics[width=\textwidth]{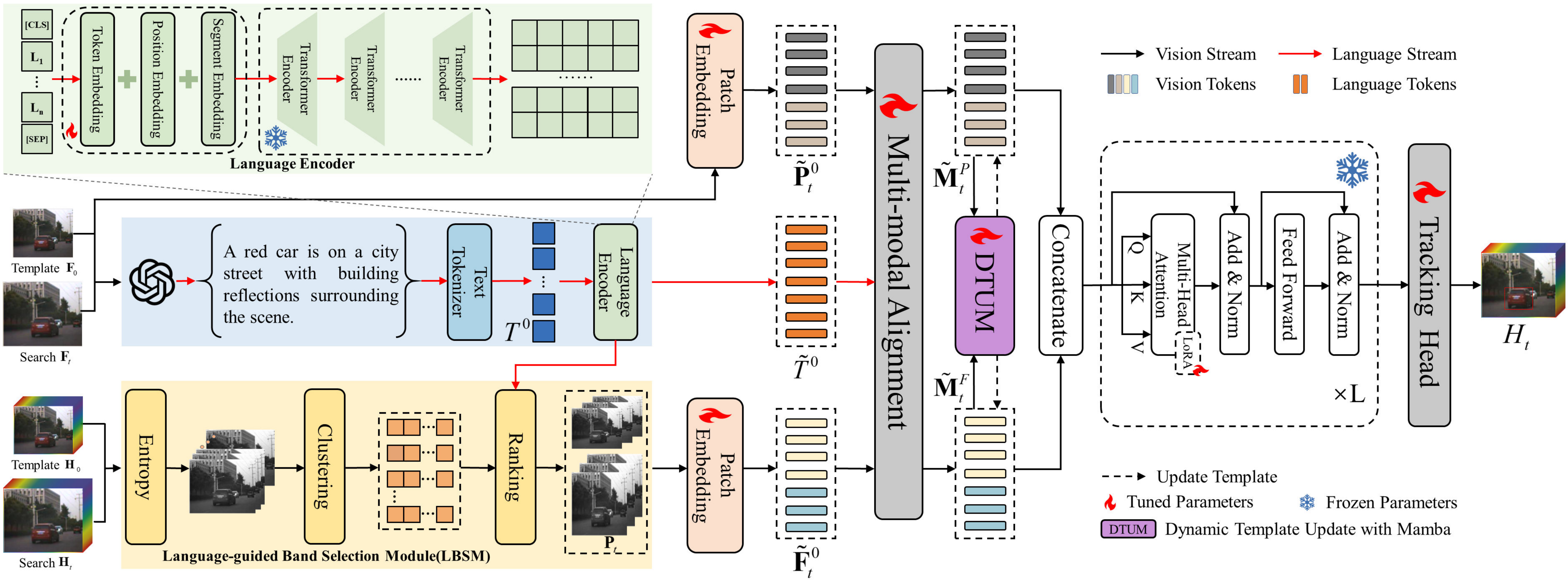} 
    \caption{{The overall architecture of VLHTrack. VLHTrack consists of LBSM, multi-modal vision–language fusion, Dynamic Template Update with Mamba (DTUM), and the prediction head for final localization. LBSM selects discriminative bands guided by structure analysis, and language semantics to form a pseudo-color image $\mathbf{P}_t$. Vision and language tokens are subsequently fused to achieve multi-modal representation learning, while DTUM models temporal dependencies and updates templates over time. The prediction head then outputs the final target state estimation.}
}
    \label{fig:VLHTrack}
\end{figure*}

\section{Method}
\subsection{Overall Architecture}
As illustrated in Fig.~\ref{fig:VLHTrack}, VLHTrack is an advanced vision–language hyperspectral tracking framework composed of three major functional modules: LBSM, multi-modal feature fusion, and DTUM, followed by a prediction head.

At each time step $t$, the network receives two complementary inputs: hyperspectral images $\mathbf{H}_t \in \mathbb{R}^{N \times H \times W}$ and false-color images $\mathbf{F}_t \in \mathbb{R}^{3 \times H \times W}$, where $N$ is the number of spectral bands, $H$ and $W$ are the height and width. $\mathbf{F}_t$ is generated from $\mathbf{H}_t$ using CIE CMFs \cite{9191105}, providing a perceptually meaningful mapping to RGB space.

{Specifically, during training and testing, we crop $\mathbf{F}_0$ using the ground-truth bounding box from the first frame, retaining only the target region and minimal context. This cropped image is fed into the LLM with a task-specific prompt: ``Please focus on the main target object in the image, and describe its color, shape, and texture in detail, within 20 words.'' 
As shown in Fig.~\ref{fig:language_description}, the LLM generates fine-grained attribute descriptions (e.g., ``reflective surface'') that naturally correlate with unique hyperspectral responses. 
It must be emphasized that this LLM functions strictly as an offline, zero-shot inference tool during initialization. This description remains fixed throughout the tracking process to prevent semantic drift from intermediate errors, ensuring that the generation process incurs zero additional training complexity.}
The language description $T$ guides band selection in LBSM and provides semantic cues for tracking, coupling spectral–physical and semantic information. 

Subsequently, the LBSM selects three representative bands from $\mathbf{H}_t$, forming the pseudo-color image $\mathbf{P}_t \in \mathbb{R}^{3 \times H \times W}$. Then $\mathbf{P}_t$, $\mathbf{H}_t$, and $T$ are input separately into the patch embedding layer and text tokenizer. These inputs are projected and flattened into a shared latent feature space, generating the corresponding token representations: 
visual tokens ($\widetilde{\mathbf{P}}^0_t$, $\widetilde{\mathbf{F}}^0_t$) and language tokens ${T}^0$. Formally, the initial token representations are given by:
\begin{equation} \label{eq:patchembed}
\begin{aligned}
    \widetilde{\mathbf{P}}^0_t &= \text{PatchEmbed}(\mathbf{P}_t), \\
    \widetilde{\mathbf{F}}^0_t &= \text{PatchEmbed}(\mathbf{F}_t), \\
    {T}^0 &= \text{Tokenizer}(\text{[CLS]}, T).
\end{aligned}
\end{equation}
Here, we denote the initial tokens generated from $\mathbf{F}_t$, $\mathbf{P}_t$ and $T$ as $\widetilde{\mathbf{F}}^0_t$, $\widetilde{\mathbf{P}}^0_t$ and ${T}^0$. 
${T}^0$ consists of $T$, [CLS] and [SEP], 
where [CLS] serves as the start token and [SEP] as the end token of the description.
{The [CLS] token represents the global semantic information of the description $T$. To leverage broad semantic priors while avoiding overfitting and preserving computational fairness, we adopt a parameter-efficient fine-tuning strategy. The deep Transformer layers of the BERT \cite{kenton2019bert} encoder are strictly frozen to retain pre-trained semantic extraction capabilities. Conversely, the shallow embedding layers (i.e., Token, Position, and Segment embeddings) are tunable, enabling the language tokens to adaptively align with the hyperspectral tracking task space. 
The text tokens $T^0$ are then processed by this language encoder to extract the semantic feature $\widetilde{T}^0$.}

These tokens are then passed through the multi-modal alignment module to achieve modality alignment. Well-aligned visual and language embeddings facilitate multi-modal representation learning and interaction. The aligned embeddings are still denoted as $\widetilde{\mathbf{M}}^p_t$ and $\widetilde{\mathbf{M}}^F_t$ for subsequent processing. 

Subsequently, we employ DTUM to update the template feature. 
Specifically, the Mamba-based DTUM module effectively captures temporal dependencies between the current search frame $\mathbf{S}_t$ and the previous template feature $\mathbf{T}_{t-1}$ (see Section~\ref{DTUM}). 
Through this mechanism, DTUM adaptively models the temporal evolution of the target’s appearance and updates the template feature through a dynamic state update process. 
Furthermore, a weighted fusion between the template feature $\mathbf{T}_{t-1}$ and the predicted feature is performed to generate the updated template feature $\mathbf{T}_t$. 

After that, $\widetilde{\mathbf{M}}^p_t$ and $\widetilde{\mathbf{M}}^F_t$ are concatenated as a sequence and fed to a $L$-layer transformer encoder. 
The concatenated embeddings are then processed within the transformer backbone, where language embeddings guide the refinement of visual features and vice versa. 
Each transformer encoder layer contains multi-head self-attention (MHSA) and a feed-forward network (FFN) with residual connections and layer normalization (LN). 
The features from the final transformer encoder layer are then used as the input to the tracking head for subsequent prediction. This bidirectional flow enables semantic-aware fusion, ensuring that the learned representations remain highly target-aware.

\subsection{Language-Guided Band Selection Module (LBSM)} \label{LBSM}

\begin{figure}[!t]
    \centering
    \includegraphics[width=\columnwidth]{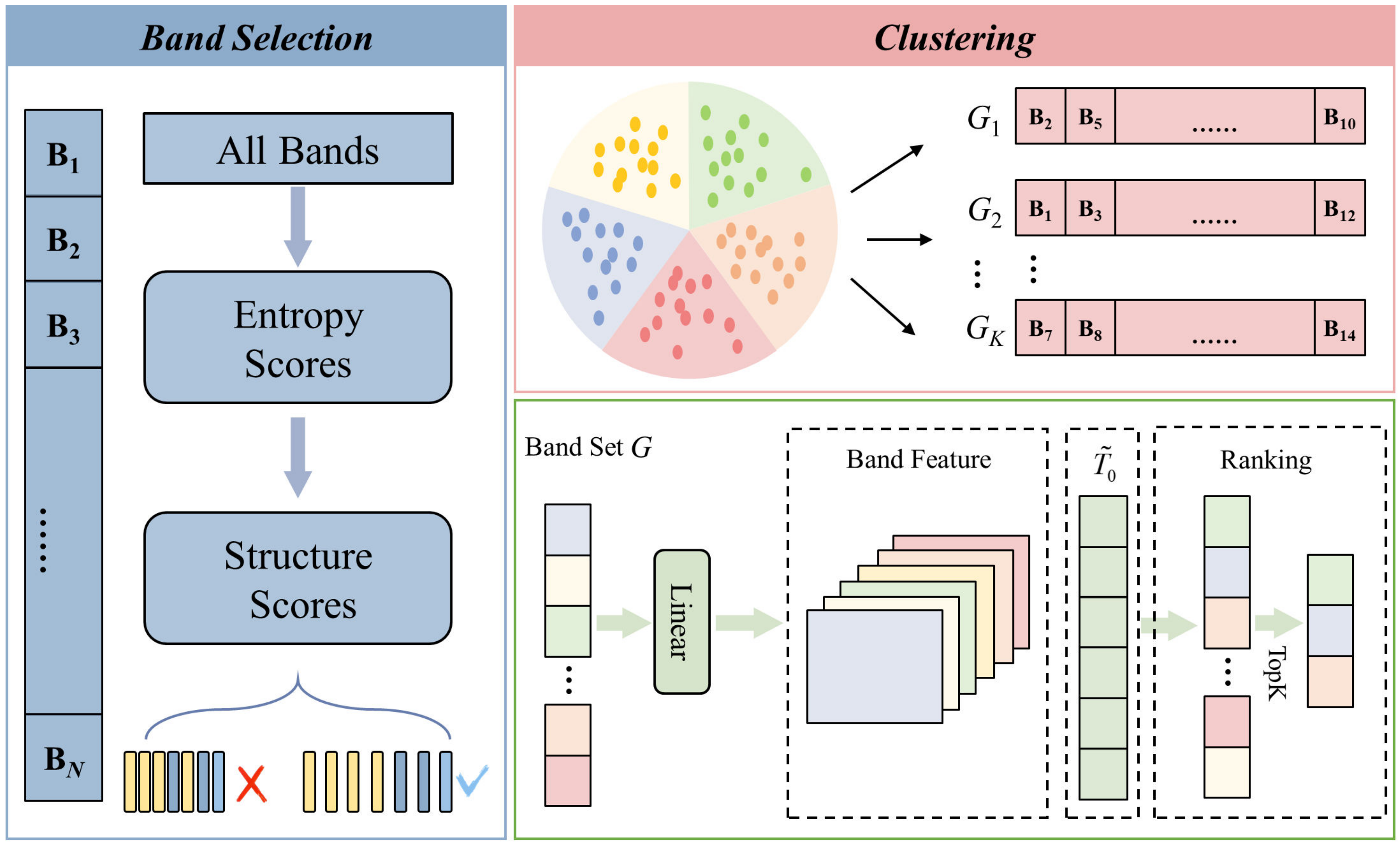} 
    \caption{{Detailed design of the proposed LBSM.}}
    \label{fig:LBSM}
\end{figure} 
 
{Hyperspectral images capture fine-grained spectral signatures that are intrinsically linked to a target's physical properties, such as material composition and surface reflectivity. However, isolating discriminative features is often hampered by severe spectral redundancy and inter-class spectral similarity. To overcome this, we conceptualize natural language as a high-level abstraction of these physical traits. For instance, the semantic prompt ``green metallic vehicle'' implicitly constrains both color reflectivity and material properties. Motivated by this inherent mapping, LBSM utilizes linguistic semantics not merely as auxiliary labels, but as physical priors. This mechanism guides the network to focus strictly on spectral bands consistent with the target's material properties, effectively filtering out spectrally similar yet semantically irrelevant background clutter.}

At time $t$, the information content of the $i$-th spectral band $\mathbf{B}_i \in \mathbb{R}^{H \times W}$ of the hyperspectral image $\mathbf{H}_t$, where $i \in \{1,\dots,N\}$, is measured by its entropy $E_i$:
\begin{equation}
E_i = -\sum_{r} p_{i,r}\log_2 p_{i,r}.
\end{equation}
Here, \(p_{i,r}\) denotes the probability density of the gray-level at pixel $r$ in  $\mathbf{B}_i$, obtained by normalizing its intensity histogram. {This normalization is performed independently within each individual band to ensure that the entropy reflects the internal statistical information of $B_i$ without being biased by global intensity variations across the spectral range.}
{Specifically, $r \in \{0, 1, \dots, 255\}$ denotes the discrete intensity levels, where the continuous spectral reflectance is quantized into 256 bins for robust entropy estimation.} A larger $E_i$ implies higher information content. 
To further suppress redundancy, each band is represented by its normalized entropy $\hat{E}_i$. {To ensure proper feature scaling and prevent the raw band index $i$ from dominating the Euclidean metric, the discrete band index is strictly normalized as $\hat{i} = {i}/{N} \in (0, 1]$.} Consequently, a pairwise distance matrix $\mathbf{D} = [D_{ij}]_{N\times N}$ is constructed, where $D_{ij}$ denotes the Euclidean distance between bands $i$ and $j$ in the $(i,\hat{E}_i)$ space. {This ensures both proximity and information content are equally weighted in density calculation.} For each band $i$, the local density $\rho_i$ and the minimum distance to higher-density bands $\delta_i$ are defined as
\begin{equation}
\rho_i = \sum_{j} \exp\!\left[-\left(\frac{D_{ij}}{\tau}\right)^2\right], \quad
\delta_i = \min_{j:\rho_j > \rho_i} D_{ij},
\end{equation}
where $\tau$ is a cutoff threshold controlling neighborhood size, determined as the 1--2\% quantile of all pairwise distances \cite{rodriguez2014clustering}, to balance locality and robustness. 
The structural redundancy score $s_i$ is then computed as
\begin{equation}
s_i = \frac{\rho_i \cdot \delta_i}{\max_k (\rho_k \cdot \delta_k)}.
\end{equation}
A higher $s_i$ indicates that the band is more representative and less redundant. 
Finally, each band is represented by a feature vector $\mathbf{v}_i = [\hat{E}_i, s_i]^\top$, jointly encoding information richness and structural distinctiveness. 
Spectral clustering is applied to $\{\mathbf{v}_i\}_{i=1}^N$, dividing all bands into $K$ clusters. 
Each cluster is defined as $G_c = \{\mathbf{B}_{u} \mid u \in \Lambda_c\}$, where $\Lambda_c$ denotes the index set of the $c$-th cluster and $c \in \{1,\dots,K\}$. 
From each cluster $G_c$, the band with the highest redundancy score $s_i$ is selected as its representative, indexed by $q = \arg\max_{i \in \Lambda_c} s_i$. 
All representatives are then aggregated to form the redundancy-suppressed candidate set $G = \{\mathbf{B}_{q} \mid q \in \Gamma\}$, where $\Gamma$ denotes the index set of the selected bands. Note that $u$ and $q$ take values in $\{1,\dots,N\}$. {It is worth noting that relying solely on heuristic statistical assumptions—such as high entropy and spectral proximity—may inadvertently retain high-entropy background noise or discard informative bands. Therefore, the preceding clustering stage serves strictly as a coarse spatial-spectral deredundancy step to ensure candidate diversity. The core of LBSM lies in this subsequent language-guided ranking. Serving as a semantic verification mechanism, it leverages physical-property-derived priors to suppress non-target high-entropy noise, rigorously assigning higher weights to bands that authentically encapsulate the target's semantic identity.}

For semantic alignment, each candidate band $\mathbf{B}_{q}$ is encoded as a low-dimensional descriptor $\mathbf{b}_{q} \in \mathbb{R}^{v}$ by a linear layer, where $v$ denotes the dimension of spectral statistical features. These are linearly projected into the shared semantic space:
\begin{equation}
\mathbf{z}_{q} = \text{Linear}(\mathbf{b}_{q}),
\end{equation}
where $\mathbf{z}_{q} \in \mathbb{R}^{V}$ and $V$ denotes the dimension of the textual embedding space. The projected feature $\mathbf{z}_{q}$ and the language embedding $\widetilde{T}^0 \in \mathbb{R}^{V}$, extracted from the language encoder, share the same dimension to ensure cross-modal consistency. 
The semantic relevance between each band and the language description is evaluated via cosine similarity:
\begin{equation} \label{eq:sim}
\text{Sim}_{q} = \frac{\mathbf{z}_{q}^{\top} \widetilde{T}^0}{\|\mathbf{z}_{q}\|\|\widetilde{T}^0\|}.
\end{equation}
Bands are ranked by $\text{Sim}_{q}$, and the top $m$ bands are retained:
\begin{equation}
S = \operatorname{TopK}\big(\{\text{Sim}_{q}: q \in \Gamma\}, m\big).
\end{equation}
In implementation, the top $m=3$ bands are fused into a three-channel pseudo-color image $\mathbf{P}_t$, preserving the most informative spectral–spatial cues while maintaining semantic coherence with the language description.

It is worth noting that the linguistic prompt remains fixed during tracking. This design leverages the physical consistency of the target's intrinsic attributes (e.g., material and spectral signatures), which are invariant to geometric deformations or localized illumination changes. A fixed semantic anchor ensures a stable spectral feature space for subsequent matching. Moreover, updating the prompt frame-by-frame would not only incur substantial computational overhead but also risk severe semantic drift by erroneously incorporating temporary occlusions or background distractors.

\subsection{Multi-Modal Vision-Language Feature Fusion}
The Multi-Modal Vision-Language Feature Fusion module aligns hyperspectral visual features with semantic language features in a unified latent space, enabling coherent cross-modal interaction. 
To align modalities efficiently, we extract only the global [CLS] token embedding, denoted as $\widetilde{T}^0$, from the encoded language sequence $T^0$ in Eqs.~\eqref{eq:patchembed}. Unlike the full sequence, which introduces linguistic noise, the [CLS] embedding compactly aggregates holistic target attributes and ensures semantic consistency with the band selection in Eq.~\eqref{eq:sim}. Subsequently, $\widetilde{T}^0$ is spatially broadcast to match the visual token layout, acting as a channel-wise semantic modulation prompt. This mechanism avoids the heavy computational overhead of cross-attention while robustly injecting language priors into the visual representation. The fusion process is formulated as:
\begin{equation} \label{eq:fusion}
\begin{aligned}
    \widetilde{\mathbf{M}}^P_t &= \widetilde{\mathbf{P}}^0_t \odot \widetilde{T}^0 + \widetilde{\mathbf{P}}^0_t, \\
    \widetilde{\mathbf{M}}^F_t &= \widetilde{\mathbf{F}}^0_t \odot \widetilde{T}^0 + \widetilde{\mathbf{F}}^0_t,
\end{aligned}
\end{equation}
where $\odot$ denotes element-wise multiplication. In this manner, the language information is seamlessly integrated into the visual embeddings through the modality alignment operation. Furthermore, since we strictly adopt the global [CLS] feature here, which is identical to that in Eq.~\eqref{eq:sim} above.
Moreover, Eqs.~\eqref{eq:fusion} \cite{zhang2023all} establish a bidirectional information flow between the visual and language modalities, thereby enabling mutual guidance for multi-modal feature extraction and interaction. By constructing such bidirectional information exchange within the Transformer backbone, the framework mitigates information dilution and enforces semantic-aware enhancement, thereby preserving strong target-awareness in the learned representations.

For the first transformer encoder layer, the fused token sequence $\widetilde{\mathbf{F}}_{\text{fusion}}$ is constructed by strictly concatenating the aligned vision-language embeddings along the sequence dimension:
\begin{equation}
\widetilde{\mathbf{F}}_{\text{fusion}} = \text{Concat}(\widetilde{\mathbf{M}}^P_t, \widetilde{\mathbf{M}}^F_t).
\end{equation}
It is then projected into the query ($\mathbf{Q}$), key ($\mathbf{K}$), and value ($\mathbf{V}$) spaces for self-attention computation:
\begin{equation}
\begin{gathered}
\mathrm{Attention}(\mathbf{Q}, \mathbf{K}, \mathbf{V}) = \mathrm{softmax}\!\left(\frac{\mathbf{Q}\mathbf{K}^\top}{\sqrt{d}}\right)\mathbf{V}, \\
\widetilde{\mathbf{E}}_t^l = \mathrm{FFN}_l\!\bigl(\mathrm{Attention}(\mathbf{Q}, \mathbf{K}, \mathbf{V})\bigr),
\end{gathered}
\end{equation}
where $\mathrm{FFN}_l(\cdot)$ denotes the feed-forward sublayer in the $l$-th transformer encoder block, and $\widetilde{\mathbf{E}}_t^l$ represents its output. For subsequent layers, the encoder takes $\widetilde{\mathbf{E}}_t^{l-1}$ as input, enabling hierarchical refinement of cross-modal representations across depth. This design effectively captures both spectral and semantic cues, ensuring stable and effective synergy between visual and language information.

\begin{figure}[!t]
    \centering
    \includegraphics[width=\columnwidth]{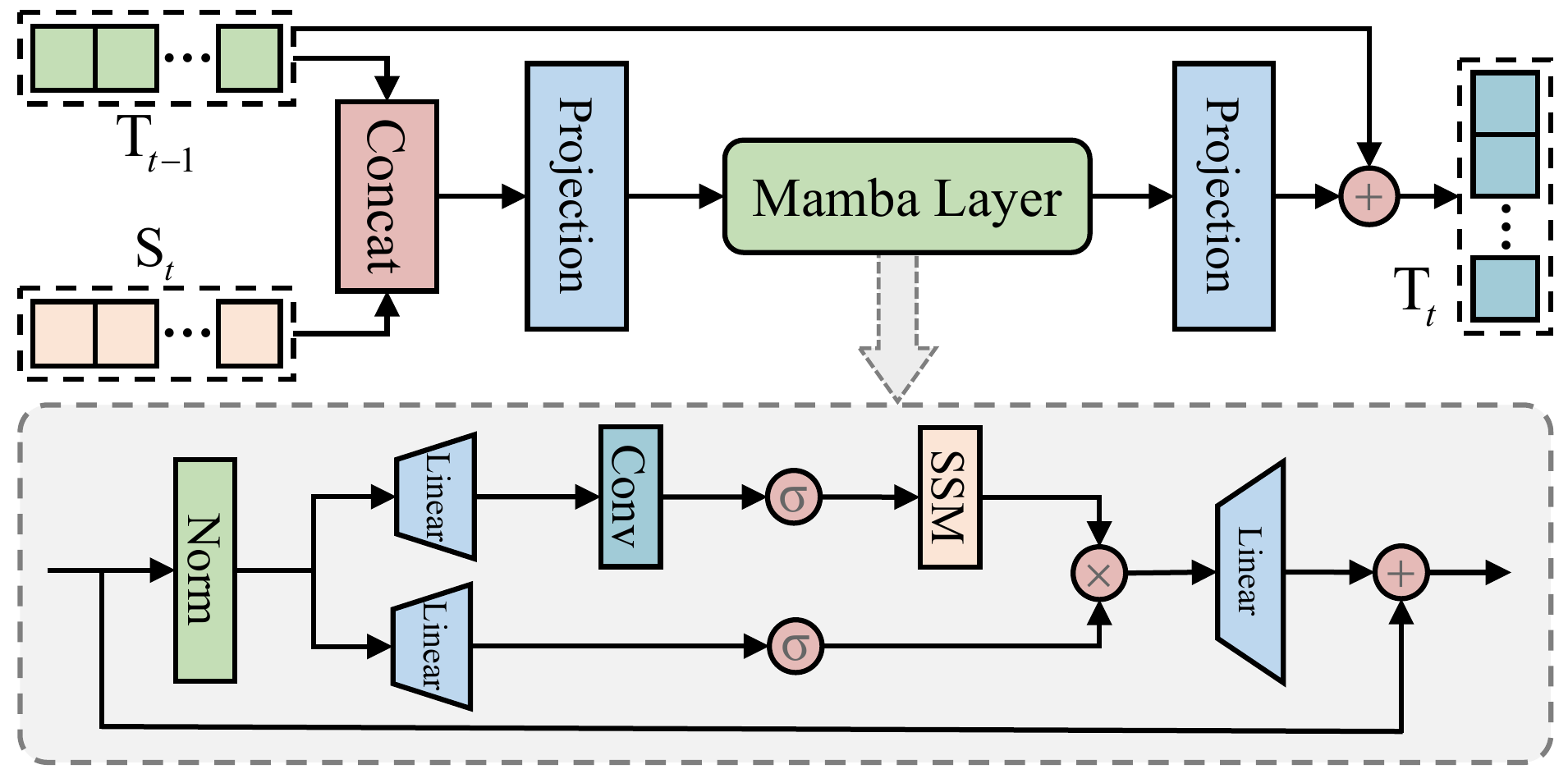} 
    \caption{Detailed design of the proposed DTUM.}
    \label{fig:DTUM}
\end{figure}

\subsection{Dynamic Template Update with Mamba (DTUM)} \label{DTUM}
To complement the static spectral prior provided by LBSM, the Dynamic Temporal Update Module (DTUM) is designed to handle dynamic challenges in long-term tracking. By leveraging Mamba's selective scanning mechanism, DTUM adaptively updates template features to capture rapid appearance variations, such as occlusions and deformations. This decoupling strategy enables VLHTrack to balance robust physical identity preservation with agile temporal adaptability. Specifically, as shown in Fig.~\ref{fig:DTUM}, DTUM is designed to model the temporal evolution of templates through selective state space modeling, thereby ensuring stable feature adaptation under drastic target appearance changes, fast motion, and long-term occlusions. 
Let the previous template feature be denoted as \(\mathbf{T}_{t-1} \in \mathbb{R}^{C \times H \times W}\) and the current search feature as \(\mathbf{S}_t \in \mathbb{R}^{C \times H \times W}\). 
Both features are then flattened into vectors: 
\begin{equation}
\begin{aligned}
\mathbf{t}_{t-1} &= \operatorname{Flatten}(\mathbf{T}_{t-1}) \in \mathbb{R}^{M}, \quad  \\
\mathbf{s}_t &= \operatorname{Flatten}(\mathbf{S}_t) \in \mathbb{R}^{M},
\end{aligned}
\end{equation}
where \(M = C \cdot H \cdot W\). These two vectors are concatenated to form a sequence:
\begin{equation}
\mathbf{X}_t = [\mathbf{t}_{t-1}, \mathbf{s}_t] \in \mathbb{R}^{2 \times M}.
\end{equation}

Direct state space modeling in the original high-dimensional space is computationally expensive and prone to overfitting. To address this, DTUM employs a learnable input projection that maps the sequence into a lower-dimensional latent space of size $M'$:
\begin{equation}
\mathbf{X}'_t = \phi(\mathbf{X}_t) \in \mathbb{R}^{2 \times M'},
\end{equation}
where \(\phi(\cdot)\) denotes the projection function implemented via linear layers with normalization and activation. Within this latent space, stacked Mamba layers are applied for temporal modeling. For each temporal token in the projected sequence $\mathbf{X}'_t = [\mathbf{x}_{t-1}, \mathbf{x}_t]$, where each $\mathbf{x}_t \in \mathbb{R}^{M'}$ represents a feature embedding at time step $t$, Mamba performs the following selective state space update:
\begin{equation}
\begin{aligned}
\mathbf{h}_t &= \mathbf{A}\mathbf{h}_{t-1} + \mathbf{B}\mathbf{x}_t, \\
\mathbf{y}_t &= \mathbf{C}\mathbf{h}_t + \mathbf{E}\mathbf{x}_t,
\end{aligned}
\end{equation}
where $\mathbf{x}_t$ is the input, \(\mathbf{h}_t\) is the hidden state, \(\mathbf{y}_t\) is the output, and \(\mathbf{A}, \mathbf{B}, \mathbf{C}, \mathbf{E}\) are learnable parameter matrices. In practice, Mamba incorporates one-dimensional convolutional filters and gating mechanisms to enhance nonlinearity and local spatial aggregation while maintaining the stability and efficiency of linear SSMs. In contrast to conventional attention mechanisms, Mamba exhibits linear complexity in sequence length, rendering it particularly amenable to long-sequence hyperspectral modeling. Stacking multiple layers further improves its ability to capture cross-frame dependencies while ensuring computational efficiency. The stacked Mamba layers process the input $\mathbf{X}'_t$ and produce the temporally modeled feature sequence, denoted as 
\begin{equation}
\mathbf{Z}_t = \text{Mamba}(\mathbf{X}'_t) \in \mathbb{R}^{2 \times M'}.
\end{equation}

After temporal modeling, the resulting sequence $\mathbf{Z}_t$ is projected back to the original flattened feature dimension through an output mapping function:
\begin{equation}
\widehat{\mathbf{Y}}_t = \psi(\mathbf{Z}_t) \in \mathbb{R}^{2 \times M},
\end{equation}
where \(\psi(\cdot)\) denotes the output mapping function. The token in $\widehat{\mathbf{Y}}_t$ is taken as the predicted updated template feature:
\begin{equation}
\widehat{\mathbf{t}}_t = \widehat{\mathbf{Y}}_t[-1, :] \in \mathbb{R}^{M}.
\end{equation}
To achieve smooth temporal adaptation, DTUM fuses the updated prediction with the previous template using a weighting strategy:
\begin{equation}
\mathbf{t}_t = \alpha \cdot \mathbf{t}_{t-1} + (1-\alpha) \cdot \widehat{\mathbf{t}}_t,
\end{equation}
where we set $\alpha$ to 0.7.
The fused vector $\mathbf{t}_t \in \mathbb{R}^{M}$ is reshaped back to the spatial domain for subsequent tracking:
\begin{equation}
\mathbf{T}_t = \operatorname{Reshape}(\mathbf{t}_t) \in \mathbb{R}^{C \times H \times W}.
\end{equation}

DTUM dynamically updates the template feature to adapt to continuous appearance variations caused by occlusions, illumination changes, and long-term motion. By employing selective state-space modeling, DTUM captures inter-frame dependencies and adaptively refines the template representation, thereby maintaining temporal coherence and robustness in dynamic tracking scenarios.

\subsection{Training of VLHTrack}
{VLHTrack is initialized with parameters from the pretrained OSTrack \cite{ye2022joint}. To enable efficient modality adaptation, we employ the Low-Rank Adaptation (LoRA) \cite{hu2022lora} technique exclusively within the vision backbone. Specifically, LoRA adapters are integrated into the output projection of the Multi-Head Attention module across all transformer blocks of the ViT. During fine-tuning, the original pretrained weights of the ViT remain completely frozen, ensuring highly parameter-efficient learning.}
The overall loss function of VLHTrack extends the base model by introducing a modality alignment loss $\mathcal{L}_{\text{vlma}}$, which enforces consistency between visual and language features. This auxiliary regularization strengthens cross-modal correspondence and improves feature coherence under spectral noise and large appearance variations.
\begin{equation}
\begin{split}
\mathcal{L}_{\text{track}} &= \mathcal{L}_{\text{cls}} + \lambda_{\text{iou}} \mathcal{L}_{\text{iou}} + \lambda_{\text{L}_\text{1}} \mathcal{L}_{1} + \mathcal{L}_{\text{vlma}},
\end{split}
\end{equation}
where $\mathcal{L}_{\text{cls}}$ denotes the weighted focal loss for classification, while $\mathcal{L}_{1}$ and generalized IoU loss $\mathcal{L}_{\text{iou}}$ are used for bounding box regression. $\mathcal{L}_{\text{vlma}}$ represents the modality alignment loss \cite{zhang2023all}. 
Following \cite{ye2022joint}, the regularization parameters $\lambda_{\text{iou}}$ and $\lambda_{\text{L}_\text{1}}$ are set to 2 and 5, respectively.

\section{Experiments}
\subsection{Experiment Settings}
\subsubsection{Dataset}
{Our primary experiments are conducted on the HOT2024 dataset from the Hyperspectral Object Tracking Challenge}%
\footnote{\url{https://www.hsitracking.com/contest}}%
{. To comprehensively evaluate the model's generalization ability, we perform supplementary experiments on the HOT2023 dataset.}
Both datasets were acquired with XIMEA snapshot hyperspectral cameras, covering VIS (16 bands), NIR (25 bands), and RedNIR (15 bands).

The HOT2023 training set consists of 110 hyperspectral video sequences, including 40 NIR, 15 RedNIR, and 55 VIS sequences. The validation set includes 87 sequences with 30 NIR, 11 RedNIR, and 46 VIS samples. Each sequence is annotated with 11 challenge attributes such as illumination and scale variation, occlusion, deformation, motion blur, rotation, out-of-view, background clutter, and low resolution, with corresponding false-color videos.

The HOT2024 dataset is an extended version of HOT2023, with significantly increased data volume and attribute coverage. It comprises a total of 217 training videos (70 NIR, 36 RedNIR and 111 VIS) and 117 validation videos (30 NIR, 20 RedNIR and 67 VIS), while adopting the same detailed attribute annotation scheme as HOT2023.

\subsubsection{Evaluation Metrics}
We adopt four standard tracking metrics: success plot, precision plot, area under the curve (AUC), and distance precision (DP). The success plot measures the proportion of frames with intersection over union (IoU) above a threshold, while AUC summarizes the overall success. The precision plot reports the proportion of frames within a distance threshold, with DP@20 pixels indicating precision at 20 pixels. All evaluations follow the One-Pass Evaluation (OPE) protocol, where the tracker is initialized on the first frame and runs through the entire sequence.

\subsubsection{Implementation Details}
The proposed method was implemented using PyTorch.
The batch size is set to 16, and the optimizer is AdamW with a weight decay of $10^{-4}$. 
The initial learning rate is set to $4\times 10^{-4}$. We trained the network for a total of 60 epochs. The backbone is initialized with a pretrained OSTrack \cite{ye2022joint} model, while all remaining trainable parameters are initialized using Xavier uniform distribution. 

\begin{figure}[t]
  \centering
  \subfloat{%
    \includegraphics[width=0.5\linewidth]{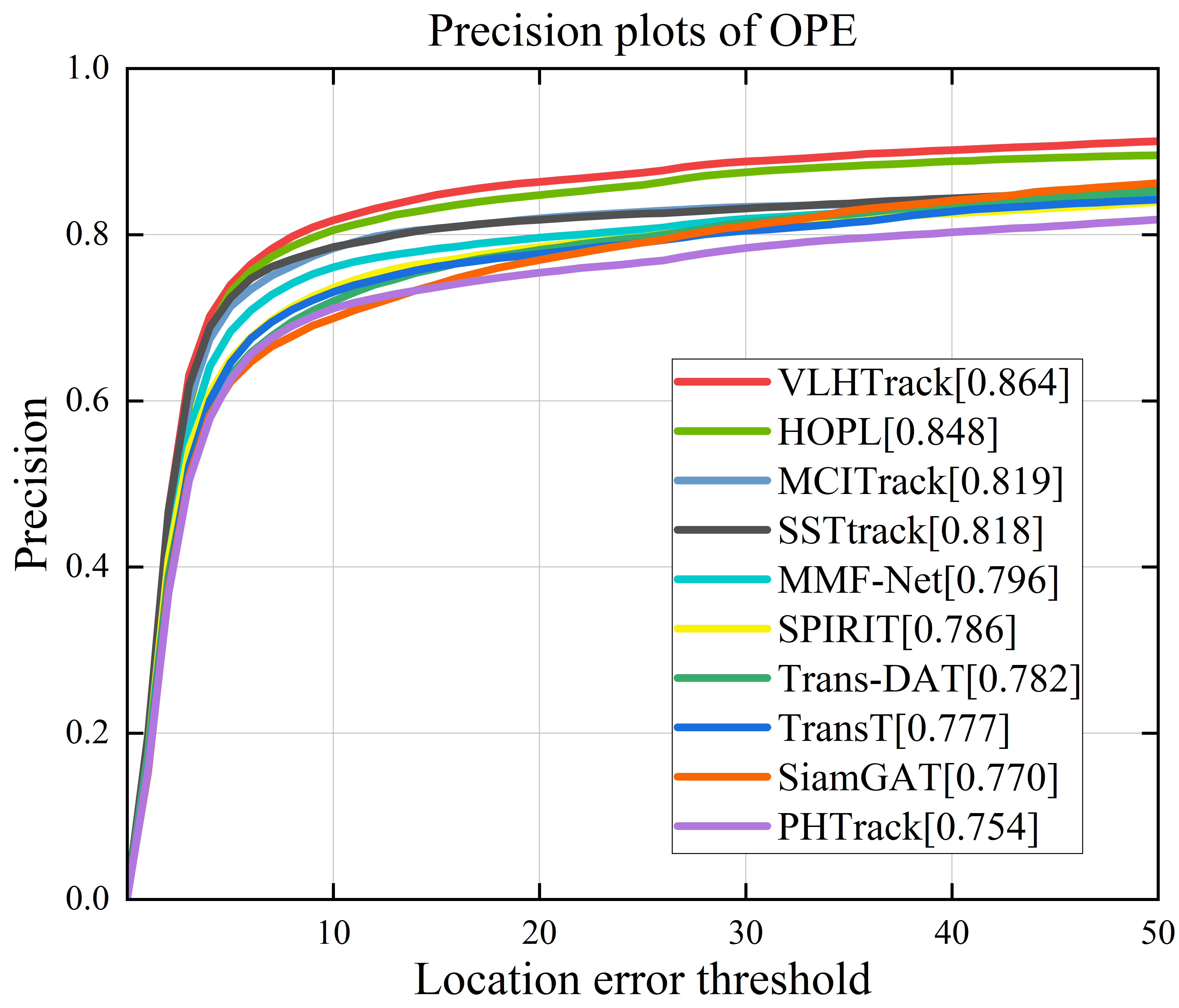}%
  }%
  \hfill
  \subfloat{%
    \includegraphics[width=0.5\linewidth]{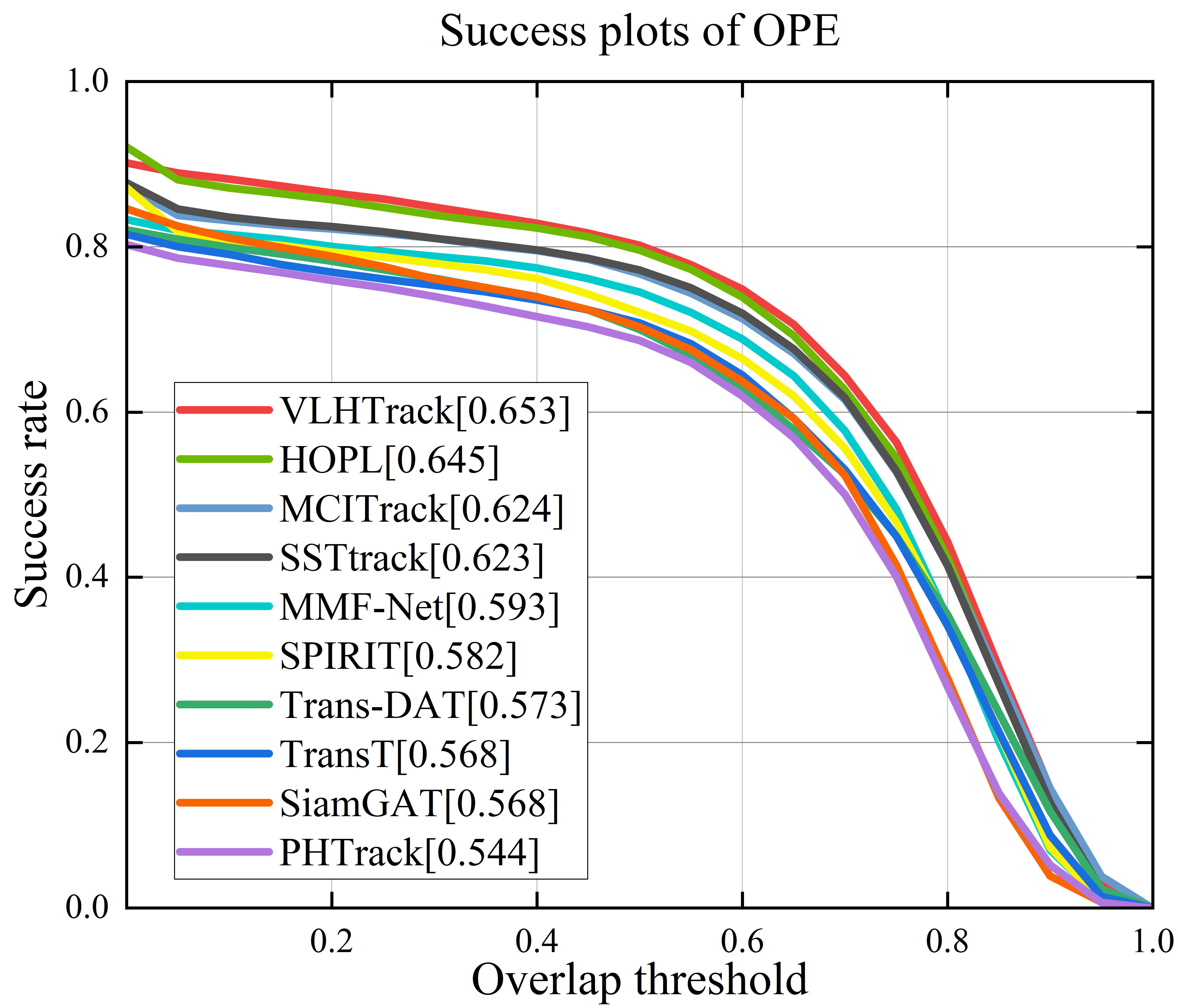}%
  }%
  \caption{Precision and success plots comparison with SOTA Trackers on the HOT2023 dataset.}
  \label{fig:HOT2023_suc&pre}
\end{figure}

\begin{figure}[t]
    \centering
    \subfloat{%
        \includegraphics[width=0.5\linewidth]{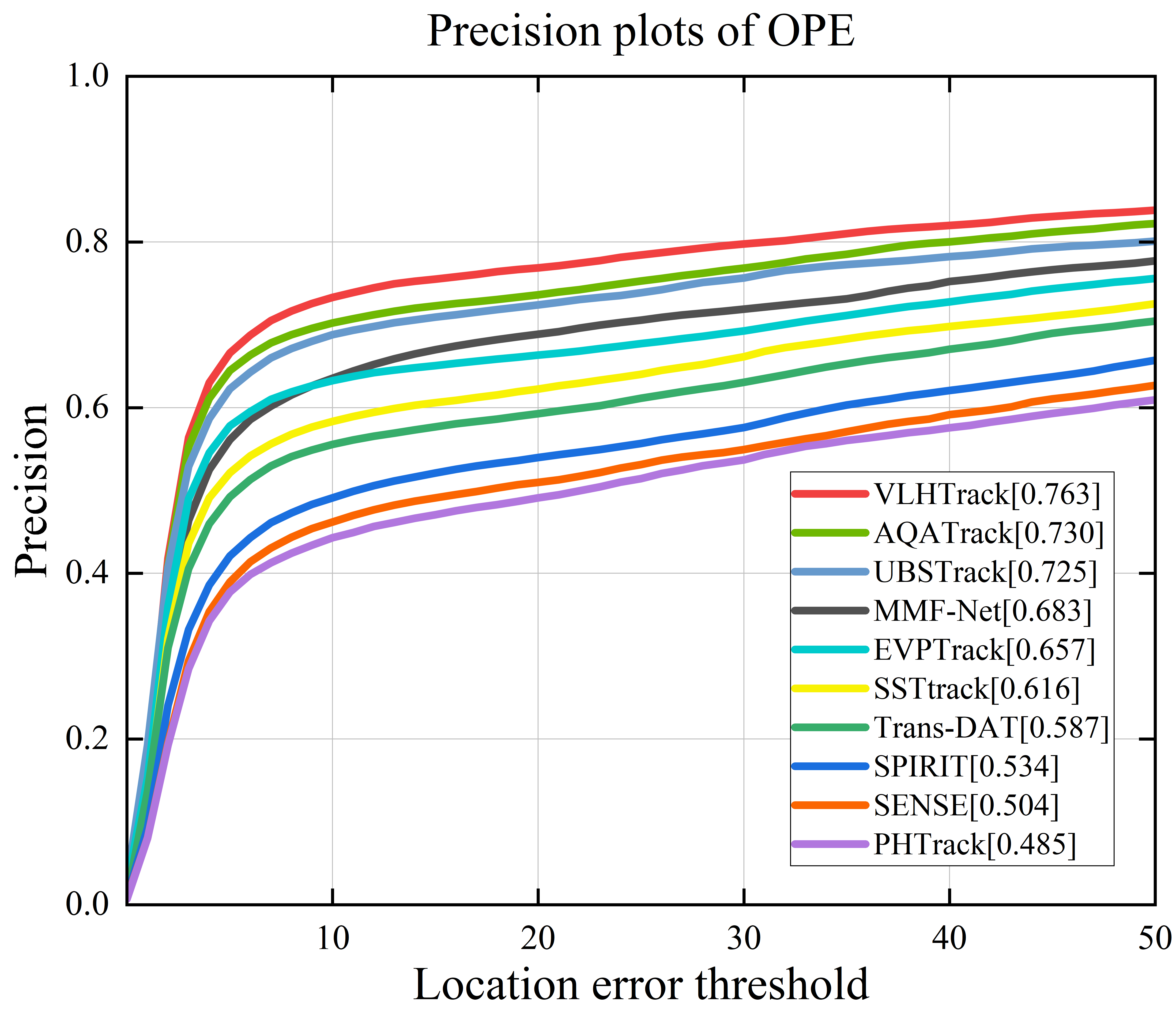}%
    }%
    \hfill
    \subfloat{%
        \includegraphics[width=0.5\linewidth]{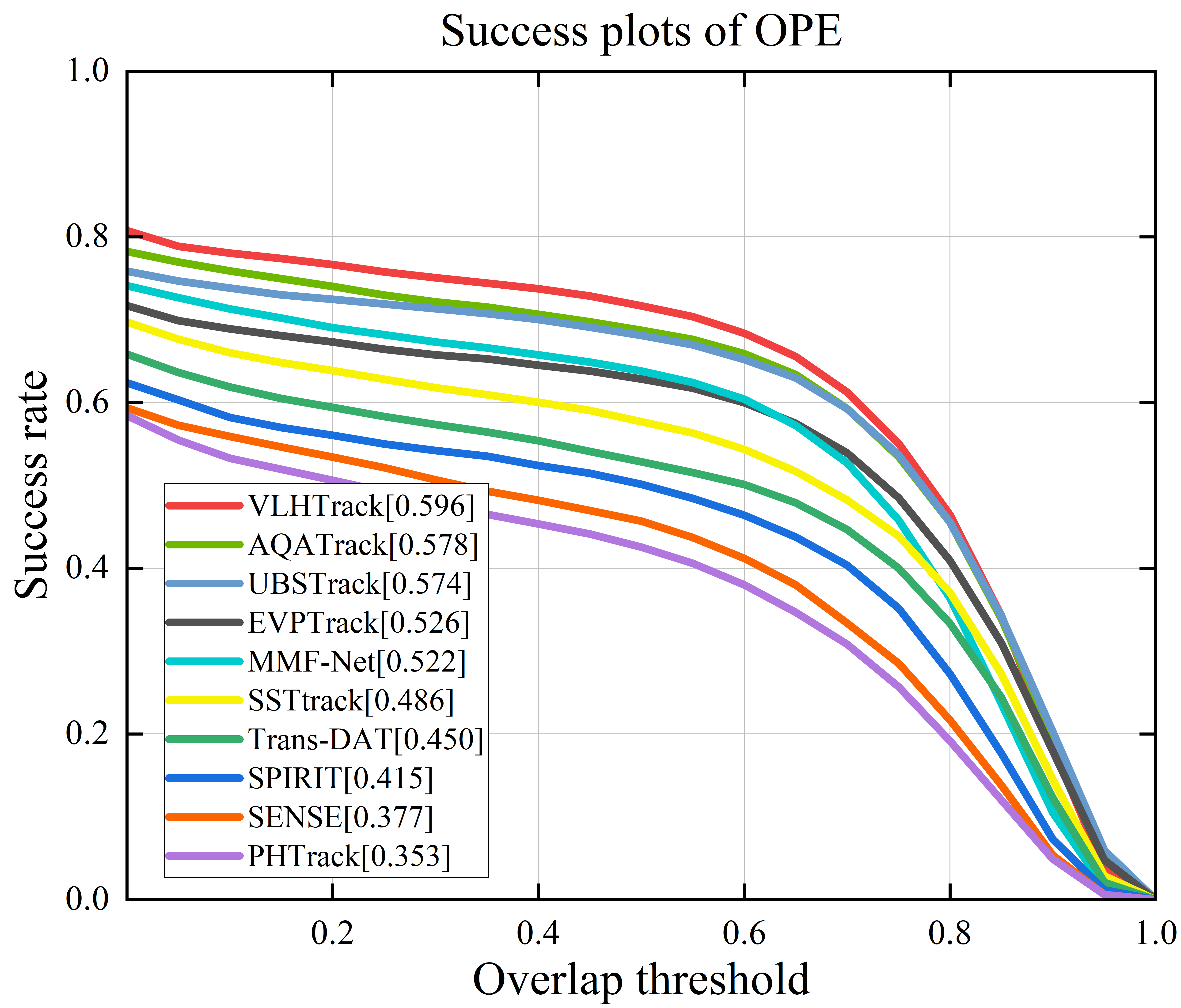}%
    }%
    \caption{Precision and success plots comparison with SOTA Trackers on the HOT2024 dataset.}
    \label{fig:HOT2024_suc&pre}
\end{figure}

\renewcommand{\arraystretch}{1.4} 
\begin{table*}[ht]
\centering
\captionsetup{justification=centering}
\caption{\protect{\textsc{Performance comparison of Vlhtrack and Sota trackers on different bands in Hot2023 dataset. The best results are marked in \textbf{bold}.}}}
\resizebox{\textwidth}{!}{%
\begin{tabular}{ccccccccccccccc}
\toprule
\multirow{2}*{\textbf{Video}} & \multirow{2}*{\begin{tabular}[c]{@{}c@{}}\textbf{Evaluation}\\ \textbf{index}\end{tabular}} & \multicolumn{3}{c}{\textbf{False-color Videos}} & \multicolumn{10}{c}{\textbf{Hyperspectral Videos}} \\
\cmidrule(lr){3-5}
\cmidrule(lr){6-15}
 &  & MCITrack & TransT & SiamGAT & HOPL & SSTtrack & MMF-Net & SPIRIT & Trans-DAT & PHTrack & {HyperTrack} & {SUIT} & {HotMoE} & VLHTrack \\
\midrule
\multirow{2}*{\begin{tabular}[c]{@{}c@{}}25 channels\\(NIR)\end{tabular}} & AUC & 0.681 & 0.595 & 0.620 & 0.725 & 0.660 & 0.666 & 0.623 & 0.681 & 0.440 & {0.738} & {-} & {-} & \textbf{0.740} \\
\cline{2-15}
 & DP & 0.870 & 0.830 & 0.858 & \textbf{0.954} & 0.854 & 0.910 & 0.838 & 0.900 & 0.558 & {0.940} & {-} & {-} & 0.942 \\
\cline{1-15}
\multirow{2}*{\begin{tabular}[c]{@{}c@{}}16 channels\\(VIS)\end{tabular}} & AUC & 0.580 & 0.634 & 0.553 & 0.641 & \textbf{0.654} & 0.612 & 0.605 & 0.512 & 0.580 & {0.643} & {-} & {0.648} & 0.621 \\
\cline{2-15}
 & DP & 0.790 & 0.847 & 0.760 & 0.854 & \textbf{0.871} & 0.815 & 0.820 & 0.721 & 0.794 & {0.864} & {-} & {0.851} & 0.850 \\
\cline{1-15}
\multirow{2}*{\begin{tabular}[c]{@{}c@{}}15 channels\\(RedNIR)\end{tabular}} & AUC & 0.443 & 0.393 & 0.449 & 0.448 & 0.395 & 0.316 & 0.371 & 0.537 & 0.412 & {0.421} & {-} & {-} & \textbf{0.552} \\
\cline{2-15}
 & DP & 0.550 & 0.527 & 0.571 & 0.558 & 0.499 & 0.410 & 0.501 & 0.715 & 0.515 & {0.583} & {-} & {-} & \textbf{0.707} \\
\cline{1-15}
\multirow{2}*{\begin{tabular}[c]{@{}c@{}}ALL channels\\(NIR, VIS, RedNIR)\end{tabular}} & AUC & 0.624 & 0.568 & 0.568 & 0.645 & 0.623 & 0.593 & 0.582 & 0.573 & 0.544 & {0.648} & {0.616} & {-} & \textbf{0.653} \\
\cline{2-15}
 & DP & 0.819 & 0.777 & 0.770 & 0.848 & 0.818 & 0.796 & 0.786 & 0.782 & 0.754 & {0.849} & {0.841} & {-} & \textbf{0.864} \\
\bottomrule
\end{tabular}
}
\label{tab:HOT2023comparison}
\end{table*}

\subsection{Quantitative Comparison with SOTA Trackers}
\subsubsection{Comparison on HOT2024 and HOT2023 Dataset}
We compare VLHTrack with representative hyperspectral and RGB-based trackers on HOT2024. Hyperspectral trackers include PHTrack \cite{10680554}, SENSE \cite{chen2024sense}, SPIRIT \cite{10375560}, Trans-DAT \cite{10491347}, SSTtrack \cite{chen2025ssttrack}, MMF-Net \cite{10438474}, and UBSTrack \cite{11007116}, and CHP \cite{11314773}, while RGB trackers such as AQATrack \cite{xie2024autoregressive} and EVPTrack \cite{shi2024explicit} are evaluated on pseudo-color sequences. Fig.~\ref{fig:HOT2024_suc&pre} presents the success and precision plots, where VLHTrack achieves the best results, highlighting its robustness and accuracy. 
We further evaluate on HOT2023 to verify generalization. As shown in Fig.~\ref{fig:HOT2023_suc&pre}, VLHTrack consistently outperforms previous methods in both success and precision, clearly confirming its adaptability to varying data distributions.

\subsubsection{Comparison on Dataset with Different Bands}
Considering the different numbers of spectral bands across modalities, we evaluate tracker performance on three spectral settings: NIR, VIS, and RedNIR. On HOT2024, nine representative hyperspectral and RGB trackers are rigorously tested on each modality. As shown in Tab.~\ref{tab:HOT2024comparison}, VLHTrack achieves the best overall performance, with an AUC of 0.596 and a DP of 0.763, while UBSTrack \cite{11007116} performs slightly better on RedNIR. This is mainly due to the limited spectral information in RedNIR, where VLHTrack’s semantic guidance becomes less effective, whereas UBSTrack’s \cite{11007116} redundancy compression favors low-dimensional spectral settings.

Similarly, on HOT2023, we compare VLHTrack with twelve methods, including HOPL \cite{zhang2025historical}, MCITrack \cite{kang2025exploring}, and SSTtrack \cite{chen2025ssttrack}, HyperTrack \cite{11131287}, SUIT \cite{xiong2025suit}, HotMoE \cite{sun2025hotmoe}, as shown in Tab.~\ref{tab:HOT2023comparison}. On HOT2023, VLHTrack achieves an AUC of 0.653 and a DP of 0.864, outperforming most competing methods. 
{Specifically, in the NIR spectral range, while VLHTrack yields a slightly lower DP score compared to HOPL \cite{zhang2025historical}, it maintains the superior AUC performance, demonstrating a better balance between tracking and overall robustness.}
However, SSTtrack \cite{chen2025ssttrack} shows superior performance on VIS sequences, likely due to
its stronger RGB-domain pretraining and resolution-aware feature modeling, which aligns better with VIS data distribution compared to the more semantically-focused VLHTrack.

Despite fierce competition in specific spectral bands—for example, SSTtrack \cite{chen2025ssttrack} performs well on HOT2023-VIS and CHP \cite{11314773} on HOT2024-RedNIR—VLHTrack demonstrates a superior overall stability. While these hyperspectral trackers excel in specific band scenarios, their performance often fluctuates when the spectral distribution changes. In contrast, VLHTrack, utilizing stable semantic anchors from frozen LLMs, achieves the highest overall score across all channels on both the HOT2023 and HOT2024 datasets. This confirms that our vision-language multi-modal model provides a more general and robust solution for practical hyperspectral tracking than methods relying solely on low-level visual features.

\begin{figure}[t]
  \centering
  \subfloat{%
    \includegraphics[width=0.5\linewidth]{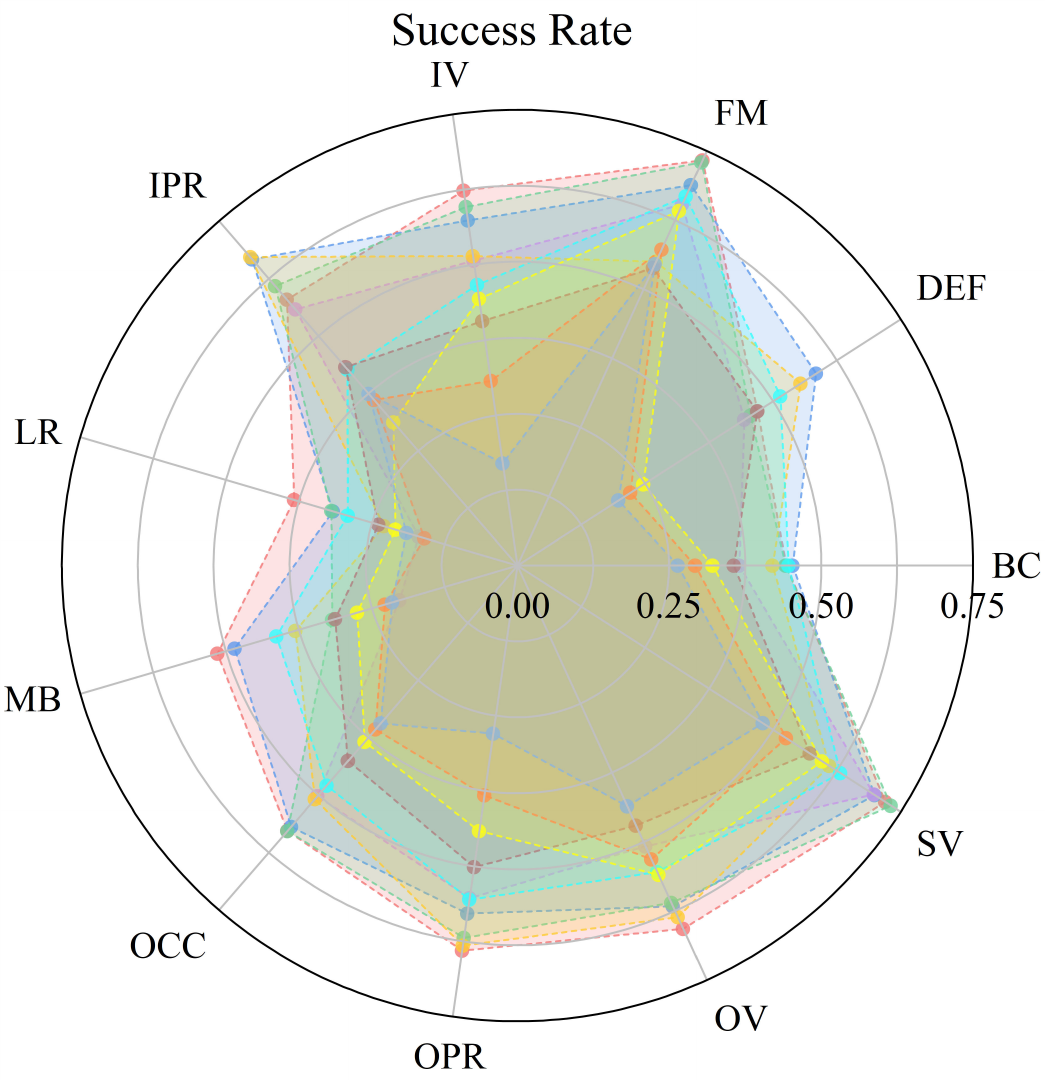}%
  }%
  \hfill
  \subfloat{%
    \includegraphics[width=0.5\linewidth]{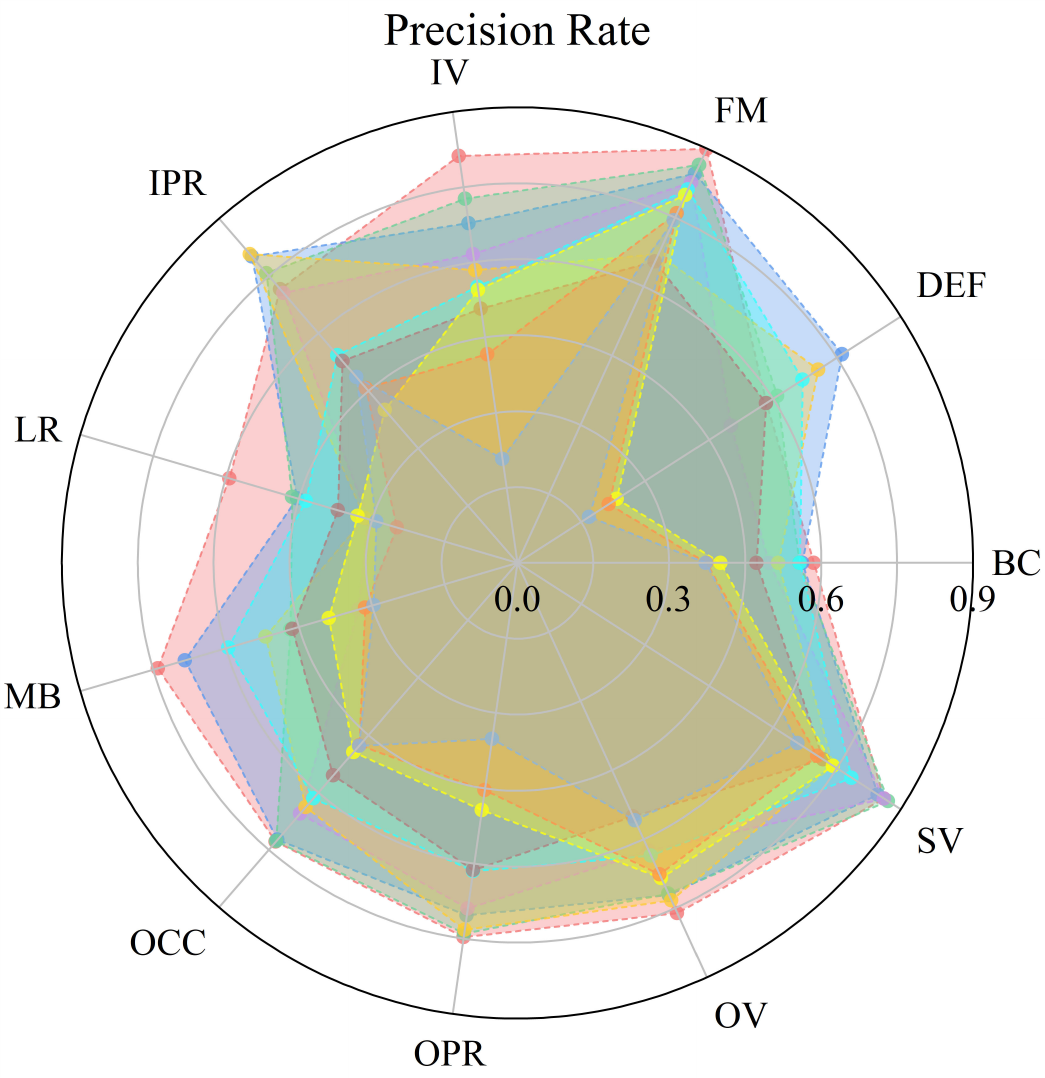}%
  }%
  
  \medskip
  \includegraphics[width=\linewidth]{image/att-legend_2024.pdf}
  \caption{The precision and success of VLHTrack and other SOTA trackers are compared on a subset of the HOT2024 dataset with 11 attributes.}
  \label{fig:att2024_suc&pre}
\end{figure}

\renewcommand{\arraystretch}{1.3} 
\begin{table*}[ht]
\centering
\captionsetup{justification=centering}
\caption{{\textsc{Performance comparison of Vlhtrack and Sota trackers on different bands in Hot2024 dataset. The best results are marked in {\textbf{bold}}.}}}
\resizebox{\textwidth}{!}{%
\begin{tabular}{cccccccccccccc}
\toprule
\multirow{2}*{\textbf{Video}} & \multirow{2}*{\begin{tabular}[c]{@{}c@{}}\textbf{Evaluation}\\ \textbf{index}\end{tabular}} & \multicolumn{2}{c}{\textbf{False-color Videos}} & \multicolumn{10}{c}{\textbf{Hyperspectral Videos}} \\
\cmidrule(lr){3-4}
\cmidrule(lr){5-14}
 &  & AQATrack & EVPTrack & UBSTrack & MMF-Net & SSTtrack & Trans-DAT & SPIRIT & SENSE & PHTrack & {HyperTrack} & {CHP} & VLHTrack \\
\midrule
\multirow{2}*{\begin{tabular}[c]{@{}c@{}}25 channels\\(NIR)\end{tabular}} & AUC & 0.712 & 0.592 & 0.725 & 0.701 & 0.712 & 0.587 & 0.655 & 0.561 & 0.527 & {0.755} & {-}  & \textbf{0.760} \\
\cline{2-14}
 & DP & 0.885 & 0.745 & 0.861 & 0.876 & 0.888 & 0.753 & 0.823 & 0.766 & 0.732 & {0.913} & {-} & \textbf{0.920} \\
\cline{1-14}
\multirow{2}*{\begin{tabular}[c]{@{}c@{}}16 channels\\(VIS)\end{tabular}} & AUC & 0.551 & 0.530 & 0.527 & 0.482 & 0.380 & 0.397 & 0.319 & 0.298 & 0.303 & {0.535} & {0.506}  & \textbf{0.557} \\
\cline{2-14}
 & DP & 0.700 & 0.658 & 0.681 & 0.645 & 0.482 & 0.524 & 0.409 & 0.398 & 0.41 & {0.709} & {0.664} & \textbf{0.735} \\
\cline{1-14}
\multirow{2}*{\begin{tabular}[c]{@{}c@{}}15 channels\\(RedNIR)\end{tabular}} & AUC & 0.471 & 0.413 & 0.486 & 0.388 & 0.504 & 0.423 & 0.377 & 0.364 & 0.262 & {0.480} & {\textbf{0.554}} & {0.478} \\
\cline{2-14}
 & DP & 0.600 & 0.521 & 0.627 & 0.521 & 0.659 & 0.547 & 0.516 & 0.465 & 0.364 & {0.621} & {\textbf{0.709}} & {0.618} \\
\cline{1-14}
\multirow{2}*{\begin{tabular}[c]{@{}c@{}}ALL channels\\(NIR, VIS, RedNIR)\end{tabular}} & AUC & 0.578 & 0.526 & 0.574 & 0.522 & 0.486 & 0.450 & 0.415 & 0.377 & 0.353 & {0.582} & {-} & \textbf{0.596} \\
\cline{2-14}

 & DP & 0.730 & 0.657 & 0.725 & 0.683 & 0.616 & 0.587 & 0.534 & 0.504 & 0.485 & {0.746}  & {-} & \textbf{0.763} \\
\bottomrule
\end{tabular}
}
\label{tab:HOT2024comparison}
\end{table*}

\begin{table*}[ht]
\centering
\caption{\textsc{Attribute-based comparison on the HOT2024 dataset. (Evaluation metric: AUC). The best results are marked in \textbf{bold}.}}
\resizebox{0.8\textwidth}{!}{%
\begin{tabular}{cccccccccccl}
\toprule
Attribute & VLHTrack & AQATrack & UBSTrack & MMF-Net & EVPTrack & SSTtrack & Trans-DAT & SPIRIT & SENSE & PHTrack\\ 
\midrule
BC & 0.443 & 0.453 & 0.441 & 0.365 & 0.420 & \textbf{0.446} & 0.356 & 0.321 & 0.293 & 0.264 \\
DEF & 0.470 & \textbf{0.584} & 0.456 & 0.444 & 0.554 & 0.514 & 0.469 & 0.246 & 0.221 & 0.198 \\
FM & \textbf{0.733} & 0.688 & 0.729 & 0.653 & 0.549 & 0.667 & 0.538 & 0.641 & 0.571 & 0.545 \\
IV & \textbf{0.623} & 0.574 & 0.596 & 0.506 & 0.514 & 0.466 & 0.406 & 0.443 & 0.307 & 0.170 \\
IPR & 0.579 & \textbf{0.666} & 0.608 & 0.558 & 0.671 & 0.427 & 0.432 & 0.312 & 0.361 & 0.374 \\
LR & \textbf{0.383} & 0.316 & 0.319 & 0.170 & 0.229 & 0.291 & 0.239 & 0.210 & 0.160 & 0.191 \\
MB & \textbf{0.514} & 0.485 & 0.317 & 0.202 & 0.381 & 0.413 & 0.312 & 0.275 & 0.227 & 0.215 \\
OCC & \textbf{0.578} & 0.569 & \textbf{0.578} & 0.502 & 0.509 & 0.479 & 0.426 & 0.384 & 0.357 & 0.344 \\
OPR & \textbf{0.640} & 0.579 & 0.619 & 0.553 & 0.632 & 0.555 & 0.501 & 0.441 & 0.382 & 0.279 \\
OV & \textbf{0.657} & 0.616 & 0.610 & 0.508 & 0.636 & 0.554 & 0.470 & 0.559 & 0.531 & 0.435 \\
SV & \textbf{0.720} & 0.699 & 0.731 & 0.697 & 0.610 & 0.632 & 0.572 & 0.596 & 0.526 & 0.480 \\
\midrule
ALL & \textbf{0.596} & 0.578 & 0.574 & 0.522 & 0.526 & 0.486 & 0.450 & 0.415 & 0.377 & 0.353 \\
\bottomrule
\end{tabular}
}
\label{tab:attribute_2024}
\end{table*}

\subsubsection{Attribute-based Comparison on the HOT2024 Dataset}
To comprehensively evaluate VLHTrack's robustness, we analyze its performance across 11 challenge attributes in the HOT2024 benchmark. Fig.~\ref{fig:att2024_suc&pre} visualizes the success and precision trends, while Tab.~\ref{tab:attribute_2024} provides the detailed quantitative AUC scores.

As shown in Tab.~\ref{tab:attribute_2024}, VLHTrack achieves the best performance in 8 out of 11 categories. Notably, in extreme scenarios like Low Resolution (LR) and Motion Blur (MB), our method secures the top rank. Specifically, in LR, VLHTrack outperforms the second-best tracker by a 20.1\% relative margin. Although absolute scores in MB are lower for all models due to spectral-spatial aliasing, VLHTrack's leading AUC demonstrates that LBSM effectively mitigates texture loss via robust semantic priors. Furthermore, for dynamic challenges like Fast Motion (FM) and Scale Variation (SV), the proposed DTUM provides crucial temporal consistency, yielding peak AUC scores of 0.733 and 0.720. However, there are performance gaps in geometry-sensitive attributes such as In-Plane Rotation (IPR) and Deformation (DEF), where VLHTrack performs slightly worse than models like AQATrack~\cite{xie2024autoregressive}. While our semantic-guided module excels at maintaining spectral-physical consistency, it is inherently less sensitive to instantaneous spatial transformations than methods utilizing specialized structural modeling or geometric augmentations. This represents a reasonable trade-off between semantic stability and geometric agility within the current vision-language fusion paradigm. Nonetheless, VLHTrack's consistent superiority across the majority of  challenge attributes solidifies its reliability for long-sequence tracking.

\begin{figure*}[htbp]
    \centering
    \includegraphics[width=\textwidth]{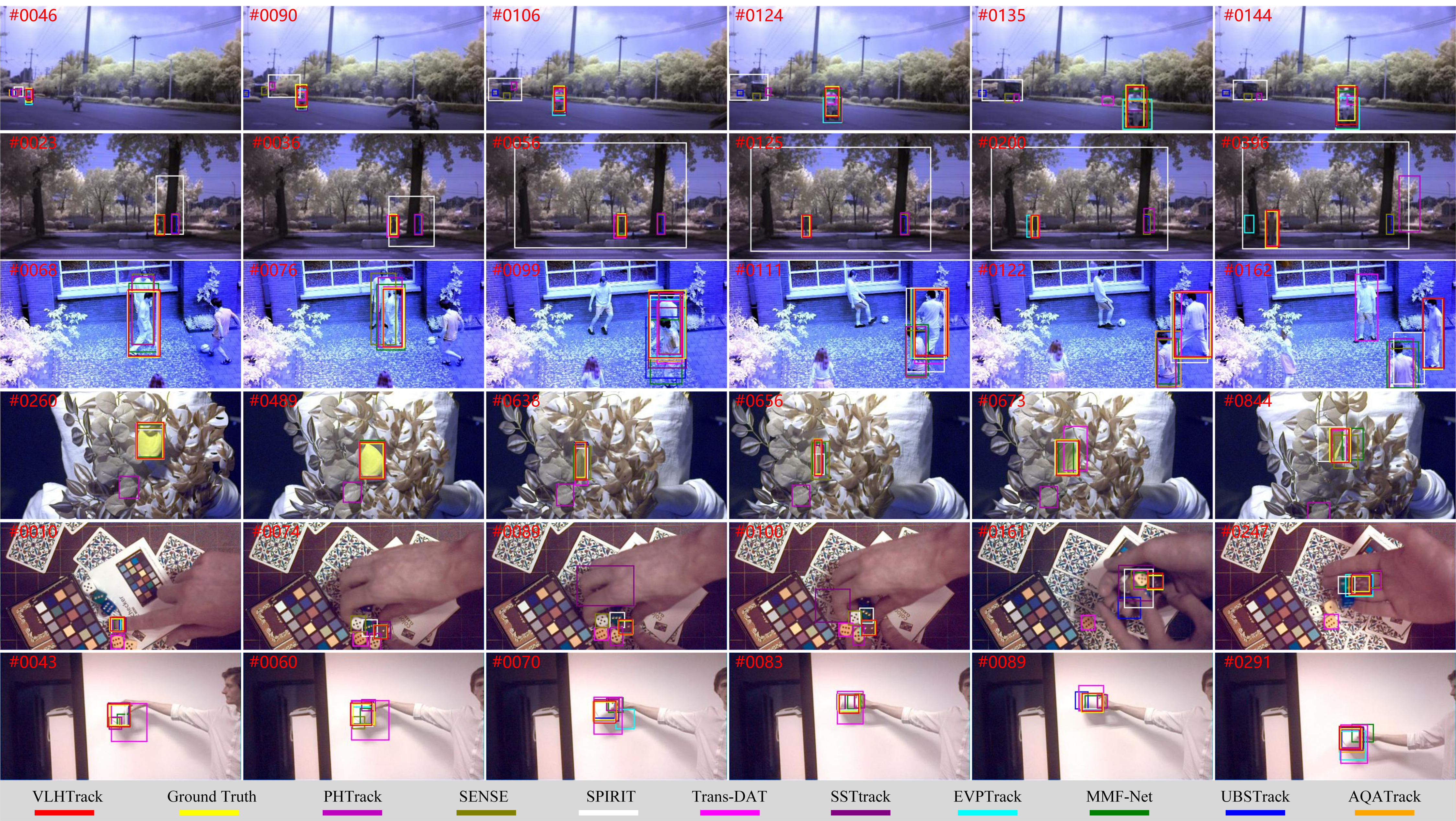}
    \caption{
    {Our proposed VLHTrack is qualitatively compared with PHTrack, SENSE, SPIRIT, Trans-DAT, SSTtrack, EVPTrack, MMF-Net, UBSTrack, AQATrack on 6 video sequences (HOT2024 dataset): \textit{nir-rider15, nir-pedestrian6, rednir-football1, rednir-leaves5, vis-dice3} and \textit{vis-whitecup2}. The proposed method performs well for these trackers. (For better illustration, we display the results in chronological order of video frames for each row, The red number in the top left corner of each image is the frame index.)}}
    \label{fig:visual-fig} 
    \vspace{-2mm}
\end{figure*}

\begin{table*}[ht]
\centering
\caption{\protect{\textsc{Performance Comparison in Terms of FPS, Params, and Flops. The Top Two Values are Bolded and Underlined.}}}
\resizebox{0.8\textwidth}{!}{%
\begin{tabular}{cccccccccc}
\toprule
\multirow{1}{*}{Videos} & \multicolumn{2}{c}{False-color Videos} & \multicolumn{7}{c}{Hyperspectral Videos} \\
\cmidrule{1-1} \cmidrule(lr){2-3} \cmidrule(lr){4-10}
Method & AQATrack & EVPTrack & UBSTrack & MMF-Net & SSTtrack & Trans-DAT & SENSE & PHTrack & VLHTrack \\
\midrule
FPS & 15.50 & 13.35 & 15.70 & 6.11 & 15.40 & \textbf{27.16} & 12.15 & 11.60 & \underline{25.50} \\
{Params(M)} & {72} & {73} & - & {\underline{45.08}} & {72.56} & {\textbf{27.98}} & {53.19} & {51.57} & {224.92} \\
{FLOPs(G)} & {58.3} & {65} & - & {352.6} & {124.95} & {\underline{30.28}} & {288.06} & {464.54} & {\textbf{23.82}} \\
\bottomrule
\end{tabular}
}
\label{tab:fps}
\end{table*}

\subsection{Qualitative Tracking Results}
To better illustrate the advantages of VLHTrack, we present qualitative comparisons in Fig.~\ref{fig:visual-fig}, involving AQATrack \cite{xie2024autoregressive}, UBSTrack \cite{11007116}, EVPTrack \cite{shi2024explicit}, MMF-Net \cite{10438474}, SSTtrack \cite{chen2025ssttrack}, Trans-DAT \cite{10491347}, SPIRIT \cite{10375560}, SENSE \cite{chen2024sense}, and PHTrack \cite{10680554}. Selected frames represent challenging scenarios with occlusion, out-of-plane rotation, and scale variation. Each row corresponds to a temporal sequence, showing prediction results over time. Ground truth boxes are shown in blue, predictions from VLHTrack in red, and other trackers in distinct colors. 
{As illustrated in Fig.~\ref{fig:visual-fig}, VLHTrack demonstrates superior robustness compared to other methods. For instance, in the \textit{nir-rider15} sequence, where the target undergoes drastic scaling amidst dense background clutter, competitive trackers like SSTtrack \cite{chen2025ssttrack} and MMF-Net \cite{10438474} experience significant drift. Conversely, VLHTrack maintains accurate localization, confirming that the semantic priors in LBSM effectively suppress background distractors. Furthermore, in the \textit{vis-whitecup2} sequence featuring high-speed motion and severe appearance distortion, purely visual trackers (e.g., AQATrack \cite{xie2024autoregressive}) completely lose the target, whereas our method remains stable.}
Overall, VLHTrack maintains stable target localization with minimal drift and distortion, particularly under occlusion or low contrast. This performance is attributed to the language-guided multi-modal fusion 
module and the context-aware template updating mechanism empowered by Mamba.
\vspace{-2mm}

\subsection{Performance Comparison of FPS, Params and FLOPs}
Tab.~\ref{tab:fps} reports the FPS, Params, and FLOPs for all evaluation methods under different video inputs. All algorithms were tested on a machine equipped with an NVIDIA RTX 2080Ti GPU. VLHTrack, MMF-Net \cite{10438474}, SSTtrack \cite{chen2025ssttrack}, Trans-DAT \cite{10491347}, SENSE \cite{chen2024sense}, and PHTrack \cite{10680554} directly process hyperspectral video input; while AQATrack \cite{xie2024autoregressive} and EVPTrack \cite{shi2024explicit} track on corresponding pseudo-color videos. As shown in Tab.~\ref{tab:fps}, VLHTrack achieves an excellent balance between tracking performance and computational efficiency. Specifically, our model achieves the lowest FLOPs among all deep learning-based hyperspectral trackers, significantly lower than models such as SENSE \cite{chen2024sense} and PHTrack \cite{10680554}. 
However, VLHTrack has a larger number of parameters compared to other methods, primarily due to the BERT language encoder \cite{kenton2019bert} and Mamba's high-dimensional projection matrices. Furthermore, since most of the parameters in the language encoder are frozen, no extra training burden is incurred. Regarding practical deployment, the LLM-generated prior operates strictly as a one-time initialization at the first frame. To quantify this overhead, our tests on the 217 hyperspectral sequences in the HOT2024 training set show that the average generation time per sequence is just 0.60 seconds, with an average of 65.54 tokens consumed per request. When amortized over long-term tracking sequences involving hundreds of frames, this initial temporal cost is negligible. Furthermore, the LLM can be immediately unloaded from GPU memory post-initialization, imposing zero computational burden during continuous tracking. This demonstrates that by introducing LBSM, we effectively bypass the computationally expensive spatial-spectral interaction modeling inherent in pure visual trackers. Furthermore, DTUM leverages Mamba’s selective scanning to avoid GPU memory bandwidth bottlenecks, maintaining a high inference rate of 25.50 FPS despite VLHTrack’s large parameter capacity. 

\begin{table}[t]
\centering
\caption{{\textsc{Ablation Study on Hot2024 Dataset: Evaluation of Vlhtrack Variants. The best results are highlighted in \textbf{bold}.}}}
\label{tab:2024_ablation}
\begin{tabular}{ccc cc}
\toprule
\multicolumn{3}{c}{\textbf{Components}} & \multicolumn{2}{c}{\textbf{Metrics}} \\
\cmidrule(r){1-3} \cmidrule(l){4-5}
\textbf{Language} & \textbf{LBSM} & \textbf{DTUM} & \textbf{AUC} & \textbf{DP@20} \\
\midrule
           &            &            & 0.580 & 0.751 \\
\checkmark &            &            & 0.584 & 0.746 \\
\checkmark & \checkmark &            & 0.590 & 0.749 \\
           &            & \checkmark & 0.592 & 0.753 \\
\checkmark &            & \checkmark & 0.592 & 0.756 \\
           & \checkmark$^{\ast}$ & \checkmark & 0.588 & 0.751 \\
\midrule
\checkmark$^{\dagger}$ & \checkmark & \checkmark & 0.593  & 0.757  \\
\checkmark & \checkmark & \checkmark & \textbf{0.596} & \textbf{0.763} \\
\bottomrule
\multicolumn{5}{l}{\footnotesize $^{\ast}$ The LBSM is incomplete as it lacks linguistic information guidance.} \\
\multicolumn{5}{l}{\footnotesize $^{\dagger}$ The entire language encoder (including embeddings) is strictly frozen.}
\end{tabular}
\end{table}

\subsection{Ablation Study}
We conduct a series of ablation studies on HOT2024 dataset to evaluate the individual and synergistic contributions of each proposed component, including the language modality, LBSM, and DTUM. To ensure a fair comparison, in all variants where our proposed LBSM is excluded, we employ the SE-Block \cite{8578843} as a standard alternative for band selection. The results are summarized in Tab.~\ref{tab:2024_ablation}. As shown in Tab.~\ref{tab:BS_ablation}, we also compare LBSM with several other band selection methods.

\subsubsection{Impact of Language Modality and LBSM}
{We first evaluate the fundamental contribution of language priors and the LBSM module. As shown in Tab.~\ref{tab:2024_ablation}, baseline yields an AUC of 0.580. 
The introduction of linguistic descriptions increased the AUC to 0.584, at which point we employed the SE-Block \cite{8578843} method for band selection. 
This gains further momentum with the introduction of LBSM, where the AUC and DP reach 0.590 and 0.749, respectively. These results indicate that semantic features effectively compensate for the limitations of pure visual representations, while LBSM successfully aligns linguistic attributes with physical spectral properties to enhance target discriminability.}

To further demonstrate the indispensability of linguistic information, we evaluate a ``purely visual'' variant that retains the advanced DTUM but removes all language-related components. In this configuration, LBSM degenerates into a basic clustering-based selection relying solely on low-level structural cues. Despite the presence of DTUM, this variant only achieves an AUC of 0.588, which is significantly lower than the 0.596 achieved by the full VLHTrack. This performance gap powerfully underscores that in complex hyperspectral scenarios, language priors are crucial for resolving target ambiguity and maintaining stable tracking.

Furthermore, we justify our language encoder fine-tuning strategy. Completely freezing the BERT \cite{kenton2019bert} encoder (including embedding layers) results in an AUC drop to 0.593 compared to the full model. This validates that tuning shallow embeddings is a parameter-efficient necessity to bridge the cross-modal gap between natural language and hyperspectral data without incurring heavy computational overhead.

\begin{table}[t]
\centering
\caption{{\textsc{Performance Comparison of Different Band Selection Methods. All variants are based on the ``Baseline + DTUM + VL Fusion'' architecture. The best results are highlighted in \textbf{bold}.}}}
\label{tab:BS_ablation}
\begin{tabular}{lcc}
\toprule
\textbf{BS Method} & \textbf{AUC} & \textbf{DP@20} \\
\midrule
Random Selection     & 0.590 & 0.755 \\
Max-Variance~\cite{7469080}        & 0.590 & 0.753 \\
SE-Block~\cite{8578843}             & 0.592 & 0.756 \\
\textbf{LBSM (Ours)} & \textbf{0.596} & \textbf{0.763} \\
\bottomrule
\end{tabular}
\end{table}

\subsubsection{Effectiveness of LBSM}
To further quantify the superiority of LBSM, We retained the language modalities and DTUM module used for tracking on top of the baseline model, and conducted comparative experiments using different band selection strategies. Specifically, we compared LBSM with three representative paradigms: Random Selection, Max-Variance \cite{7469080}, and SE-Block \cite{8578843}. In hyperspectral images, bands that are far apart in the spectral sequence tend to have weak correlations. We used a random selection method to select bands that are far apart in the spectrum (e.g., bands 1, 5, and 15) to generate pseudo-color representations. As shown in Tab.~\ref{tab:2024_ablation}, when we replaced LBSM with a random band selection strategy, the AUC reached only 0.590 despite retaining the language modality and DTUM. Compared to the full VLHTrack model, its AUC and DP metrics decreased by 0.6\% and 0.8\%, respectively.
Specifically, for Max-Variance \cite{7469080}, we rank the channels based on their independent spatial variance and extract the top 3 bands to construct the false-color input. For the data-driven SE-Block \cite{8578843}, we retain the top 3 bands corresponding to the highest learned attention weights.
As shown in Tab.~\ref{tab:BS_ablation}, while Max-Variance \cite{7469080} and SE-Block \cite{8578843} outperform random selection by utilizing statistical entropy and visual attention respectively, both fall short of LBSM. 
Max-Variance \cite{7469080} often excessively retains high-variance background noise, and SE-Block, lacking physical constraints, is prone to overfitting spectral shortcuts, making it difficult to distinguish the target from salient yet semantically irrelevant clutter. In contrast, by leveraging linguistic knowledge as supervision, LBSM effectively addresses high-dimensional redundancy and achieves a significantly more robust physical-semantic alignment than implicit, self-supervised weighting.
These results demonstrate that LBSM effectively aligns spectral channels with semantic relevance, thereby reducing redundant spectral information.

\begin{table}[t]
\centering
\caption{{\textsc{Performance Comparison of Different Template Update Mechanisms. All variants are based on the ``Baseline + LBSM + VL Fusion'' architecture. The best tracking performance (AUC, DP@20, FPS) and the minimal Params/FLOPs are highlighted in \textbf{bold}.}}}
\label{tab:DTUM_ablation}
\resizebox{\linewidth}{!}{%
\begin{tabular}{lccccc}
\toprule
\textbf{Mechanism} & \textbf{AUC} & \textbf{DP@20} & {\textbf{Params(M)}} & {\textbf{FLOPs(G)}} & \textbf{FPS} \\
\midrule
Linear Weighted & 0.588 & 0.752 & {\textbf{167.79}} & {\textbf{22.91}} & 21.40 \\
{LSTM} & {0.583} & {0.743} & {228.40} & {24.59} & {19.80} \\
{Transformer} & {0.587} & {0.750} & {235.96} & {25.83} & {18.60}\\
\textbf{DTUM (Ours)} & \textbf{0.596} & \textbf{0.763} & {224.92} & {23.82} & \textbf{25.50}  \\
\bottomrule
\end{tabular}%
}
\end{table}

\subsubsection{Role of DTUM}  
We further evaluated the contribution of DTUM. As shown in Tab.~\ref{tab:2024_ablation},  
after incorporating DTUM into the model based on LBSM and language modality guided tracking, the AUC improved from 0.590 to 0.596, and the DP increased from 0.749 to 0.763. Secondly, even in configurations employing weaker SE-Block \cite{8578843}, adding DTUM still delivers stable performance gains: AUC increased from 0.584 to 0.592, and DP rose from 0.746 to 0.756. This underscores the necessity of Mamba-based temporal context modeling for capturing the evolution of target appearance. 

Furthermore, to validate DTUM's superiority, we compared it against LSTM and Transformer variants under identical settings, as shown in Tab.~\ref{tab:DTUM_ablation}. While traditional linear weighting is computationally efficient, its limited capacity for non-linear modeling yields a subpar AUC of 0.588. Transformer-based modeling, despite its global receptive field, suffers from quadratic complexity $O(L^2)$, which leads to an increase in FLOPs to 25.83G and a reduction in inference speed to 18.60 FPS. Notably, both LSTM and Transformer variants underperform even the standard linear weighted template update method in AUC, likely due to template contamination where self-attention mechanisms aggregate irrelevant background noise. 
In contrast, DTUM achieves a peak AUC of 0.596 and 25.50 FPS. Remarkably, although DTUM has higher parameters than standard linear weighted template update method, it is significantly faster. This efficiency stems from Mamba’s hardware-aware selective scan mechanism, which optimizes GPU utilization by performing SRAM-resident computations and mitigating memory bandwidth bottlenecks. By balancing $O(L)$ complexity with input-dependent gating, DTUM effectively preserves temporal consistency without sacrificing real-time performance.

\section{Conclusion}
We present VLHTrack, a vision–language hyperspectral tracking framework that effectively integrates natural language into HOT. The proposed LBSM establishes a semantic-to-spectral mapping that aligns linguistic semantics with spectral–physical properties, thereby enhancing feature discriminability and reducing spectral redundancy. Meanwhile, DTUM leverages selective state space modeling to capture inter-frame dependencies and update templates under dynamic conditions. Extensive experiments on the HOT2023 and HOT2024 datasets demonstrate that VLHTrack achieves superior performance and strong generalization capability.

\bibliographystyle{IEEEtran}
\bibliography{sample}

\end{document}